%% file: submission_main.tex
\documentclass[pdflatex,sn-basic]{sn-jnl}

\usepackage{amsmath,amssymb,amsfonts}
\usepackage{graphicx}
\usepackage{textcomp}
\usepackage{booktabs}
\usepackage{tabularx}
\usepackage{bm}
\usepackage{url}
\usepackage{microtype}
\usepackage{flafter}
\usepackage{placeins}

\newcommand{\PARstart}[2]{#1#2}
\newcommand{\fullwidthgraphic}[2][\linewidth]{\includegraphics[width=#1]{#2}}

\makeatletter
\renewcommand{\keywords}[1]{\def\@keywords{\par\addvspace{10pt}{\keywordfont{\bfseries\keywordname:} #1\par}}}
\makeatother

\begin{document}

\title[METER for short-event microstate transitions]{METER: Matched-Event Calibration for Short-Event EEG Microstate Transition Analysis}

\author*[1]{\fnm{Abdul} \sur{Basit}}\email{abdul.basit@nyu.edu}
\equalcont{These authors contributed equally to this work.}
\author[2]{\fnm{Eklavya} \sur{Tomar}}
\equalcont{These authors contributed equally to this work.}
\author[1]{\fnm{Muhammad} \sur{Shafique}}

\affil[1]{\orgname{eBRAIN Lab, New York University Abu Dhabi}, \orgaddress{\city{Abu Dhabi}, \\ \country{United Arab Emirates}}}

\affil[2]{\orgname{Dubai College}, \orgaddress{\city{Dubai}, \country{United Arab Emirates}}}

\abstract{\input{Sections/01_abstract}}


\maketitle

\input{Sections/02_introduction}

\input{Sections/03_background}
\input{Sections/04_methodology}
\input{Sections/05_experimental_setup}
\input{Sections/06_results}
\input{Sections/07_conclusion}

\bmhead{Acknowledgments}
The authors thank the creators and maintainers of the ZuCo 2.0 corpus for making the physiological recordings and annotations publicly available.

\section*{Supplementary information}
\noindent\textbf{Supplementary Material:} Representation and preprocessing validation, complete estimator and calibration implementation, comparison-method specifications, group and application analyses, and operating-characteristic robustness evidence.

\section*{Statements and Declarations}
\noindent\textbf{Funding.} No funding was received for conducting this study.

\noindent\textbf{Competing interests:} The authors declare no competing financial or non-financial interests relevant to this work.

\noindent\textbf{Ethics approval:} This study is a secondary analysis of publicly released, de-identified ZuCo 2.0 data and involved no new participant recruitment, intervention, or access to identifiable information. The original collection was approved by the Ethics Commission of the University of Zurich, as documented in the source publication \citep{r3}.

\noindent\textbf{Consent to participate:} The original dataset investigators report that participants provided written informed consent for participation and reuse of the released data \citep{r3}.

\noindent\textbf{Consent for publication:} Not applicable.

\noindent\textbf{Data availability:} The ZuCo 2.0 dataset is publicly available from the Open Science Framework \citep{r20}.

\noindent\textbf{Code availability:} The analysis code, versioned analysis configurations, and dependency specifications are available from the corresponding author on request. A versioned release will be deposited in a public repository with a persistent identifier before publication.

\noindent\textbf{Author contributions:} E.T. conceived the extension of microstate transition analysis to naturalistic reading, built the original analysis pipeline including sentence-level EEG reconstruction, microstate segmentation, and the optimal transport implementation, identified and characterized the finite-event artifacts in per-sentence transition costs including the role of unvisited-state rows, performed the cross-participant map alignment analysis, and produced the archived analysis used for implementation validation. A.B. developed the METER framework, estimators, and calibration procedure, performed the simulation and ablation studies, expanded the dataset reconstruction to the full cohort, and drafted the manuscript. E.T. and A.B. jointly revised the manuscript. M.S. supervised the research. All authors reviewed and approved the final manuscript.

\noindent\textbf{AI Disclosure:} OpenAI ChatGPT was used for code refinement and grammar review. The authors verified all suggested changes and remain fully responsible for the manuscript content.

\bibliography{references_springer}

\end{document}


\begin{center}
{\Large\bfseries Supplementary Material}\\[6pt]
{\large METER: Matched-Event Calibration for Short-Event EEG Microstate Transition Analysis}
\\[7pt]
Abdul Basit$^{1,\dagger}$, Eklavya Tomar$^{2,\dagger}$, and Muhammad Shafique$^{1}$\\[3pt]
$^{1}$\,eBRAIN Lab, New York University Abu Dhabi, Abu Dhabi, United Arab Emirates\\
$^{2}$\,Dubai College, Dubai, United Arab Emirates\\
$^{\dagger}$These authors contributed equally to this work.\\[3pt]
Corresponding author: Abdul Basit, \texttt{abdul.basit@nyu.edu}\\[3pt]
Journal: \textit{Brain Topography}
\end{center}

\input{Sections/08_supplement}

%% file: Sections/01_abstract.tex
Short cognitive events contain unequal and often sparse transition evidence, so raw EEG microstate distances can reflect finite sampling rather than neural reconfiguration. We introduce \emph{METER} (\underline{M}atched-\underline{E}vent \underline{T}ransition \underline{E}stimation against a \underline{R}eference), a participant-relative framework that separates marginal occupancy from conditional next-state choice and evaluates each event against duration-conditioned events from a fixed participant-specific reference bank. METER-Jump integrates uncertainty in sparse target and reference transition tables, returning a randomized event rank for testing and a continuous standardized coordinate for association analysis. Using a reproducible three-state representation selected by spatial criteria, we evaluated METER with participant-fitted simulations, estimator ablations, map perturbations, and preprocessing sensitivities. Event-level false-positive rates were 0.044--0.056 across four heterogeneous null mechanisms. By contrast, global cutoffs learned under the participant-fitted null and transferred unchanged to other regimes produced false-positive rates from 0.004 under rare transitions to 0.234 under map perturbation; duration-conditioned ranks remained near 0.05. Detection probability was approximately 0.69 with fixed maps but fell to 0.22 after bootstrap-derived map-assignment perturbation. In 4,175 natural-reading events from 18 participants, no linguistic association survived multiplicity correction in the primary 30-ms analysis, although a 20-ms sensitivity changed one within-stream max-$T$ decision. METER therefore calibrates finite-event departures conditional on a defined microstate sequence, while map assignment, filtering, temporal smoothing, and reference construction remain separate sources of measurement uncertainty.

%% file: Sections/02_introduction.tex
\section{Introduction}
\label{sec:introduction}
\PARstart{E}{EG} microstates describe ongoing brain activity as a succession of brief, quasi-stable scalp topographies. Their duration, occurrence, and coverage can be reproducible, but transition probabilities are markedly less reliable: a large retest study found mostly poor-to-moderate transition reliability despite good reliability for simpler temporal characteristics \citep{r40}. This distinction matters whenever microstate dynamics are examined within individual cognitive events. Event-level analysis could reveal how whole-scalp dynamics change from one sentence, trial, or behavioral episode to the next, yet each event provides only a short and uneven sample of the underlying transition process.

\begin{figure*}[!h]
\centering
\fullwidthgraphic{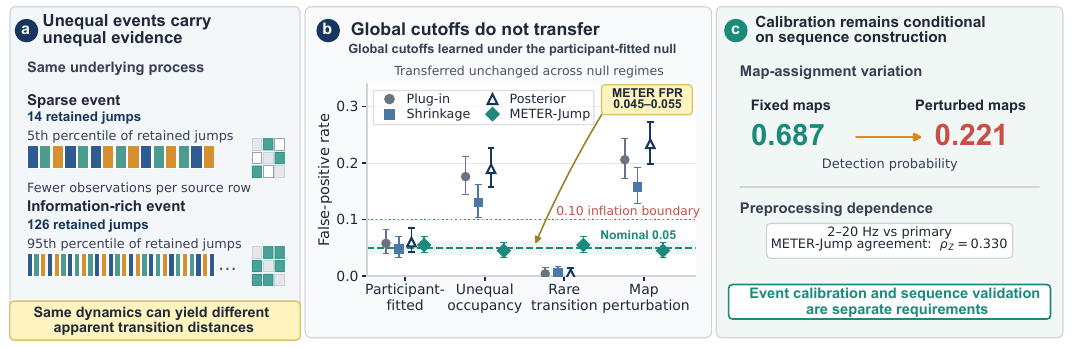}
\caption{Motivation and operating boundary of short-event transition calibration. (a) Events generated by the same underlying process can contain sharply different amounts of transition evidence, so sparse conditional rows can produce larger apparent distances. (b) Fixed global cutoffs transfer poorly across participant-fitted null regimes, whereas duration-conditioned METER calibration remains near the nominal false-positive rate (0.045--0.055) in the paired estimator-ablation simulations. The separately seeded evaluation simulations are reported in Fig.~\ref{fig:calibration}. (c) Event calibration does not remove uncertainty introduced before scoring: map perturbation reduces detection probability, and coordinate agreement under an alternative passband falls to $\rho_Z=0.330$. Finite-event calibration and sequence validation are therefore complementary rather than interchangeable requirements}
\label{fig:motivation}
\end{figure*}

Figure~\ref{fig:motivation} summarizes the three measurement problems that motivate METER. Events with the same dynamics can carry unequal transition evidence (panel a), cutoffs learned under one sampling regime need not transfer to another (panel b), and any calibrated event test remains conditional on the sequence produced by map assignment and preprocessing (panel c). METER addresses finite-event variation while sequence reliability remains a separate requirement.

Naturalistic reading makes this measurement problem concrete. Sentences vary substantially in duration and consequently contain different numbers of realized microstate jumps. Consider two sentences generated by the same transition process, one with only a few observed jumps and the other with many. The shorter sequence produces sparser conditional rows and can appear farther from a reference simply because it carries less transition evidence. Unequal state prevalence compounds the problem: frequently visited states yield well-supported destination estimates, whereas rare states may contribute nearly empty rows. A raw event-level transition distance can therefore reflect how much information happened to be observed, and how that information was distributed across states, rather than a genuine change in neural dynamics.

The scientific interpretation is further complicated when different mechanisms are collapsed into one score. A joint transition distribution can change because the marginal prevalence of a state changes even when the next-state choices conditional on that state remain constant. Conversely, destination choices can be reorganized while marginal prevalence changes little. Occupancy and conditional switching are both meaningful, but they answer different questions. A method intended to characterize event-level reconfiguration should distinguish them before assigning biological meaning to a transition effect.

Finite-event uncertainty is also separate from uncertainty in the state sequence itself. Microstate maps must first be assigned to continuous EEG, and ambiguity between similar topographies can turn modest spatial variation into different discrete labels. Standardized workflows, objective template comparison, and explicit preprocessing choices are consequently central to microstate research \citep{r41,r42,r43,r45}. Reliable group maps do not guarantee reliable jump sequences, and downstream calibration cannot reconstruct transitions that filtering, map assignment, or temporal smoothing has removed or reassigned. Event-level inference therefore requires two linked but distinct questions: whether the sequence is sufficiently stable to interpret, and whether an observed deviation is unusual given the finite information within that sequence.

This study addresses the second question while evaluating the first separately. Specifically, we ask: \emph{does one short, already labeled microstate event differ in its first-order destination choices from a fixed participant-specific reference by more than its finite transition evidence would predict?} Conventional summaries describe observed temporal characteristics \citep{r1,r5}; entropy and syntax measures characterize sequence organization \citep{r44,r46}; hidden Markov and semi-Markov models estimate latent dynamics and duration laws \citep{r29,r30,r34}; and endpoint transport quantifies marginal redistribution \citep{r2}. These methods answer adjacent questions, but none supplies, by default, the target-disjoint, participant-specific finite-event test posed here.

We introduce \textit{METER} (\underline{M}atched-\underline{E}vent \underline{T}ransition \underline{E}stimation against a \underline{R}eference) to fill this gap. In plain terms, METER asks how different an event's next-state choices are from that participant's usual pattern, relative to comparable events. \textit{METER-Jump} isolates conditional jump-destination reconfiguration, whereas \textit{METER-Occ} separately reports state-prevalence change. Posterior integration accounts for sparse target and reference tables, and matched pseudo-events calibrate each target against a participant-specific bank fixed without linguistic outcomes. The target never updates that bank. Dwell reconfiguration remains exploratory rather than entering the primary transition claim.

The ZuCo 2.0 Natural Reading corpus provides a stringent test because sentence duration, transition evidence, and linguistic properties vary together \citep{r20,r3}. We evaluate METER in participant-fitted simulations, quantify the effects of map and preprocessing uncertainty, and apply it to 4,175 natural-reading events from 18 participants. This design separates calibration of the event test from reliability of the state sequence on which that test operates.

\subsection{Contributions and scope}

The study makes four novel contributions:
\begin{enumerate}
\item \textbf{An operationally separated event-level estimand:} METER distinguishes marginal state prevalence from first-order conditional jump destinations, giving each effect a clear interpretation while keeping censor-aware dwell change explicitly exploratory.

\item \textbf{Participant-relative finite-event calibration:} A fixed, target-disjoint participant-specific reference is combined with uncertainty-aware transition estimation and duration-matched pseudo-events. The result is both a randomized event rank for testing and a continuous standardized coordinate for association analysis, without relying on a single global rejection cutoff.

\item \textbf{Separate validation of event calibration and sequence reliability:} Participant-fitted simulations and targeted ablations distinguish control of finite-event error from map reliability, sequence reliability, and downstream effect stability. They show why regularization alone is insufficient and why nominal calibration cannot establish measurement invariance.

\item \textbf{Naturalistic evaluation under heterogeneous event information:} We compare the proposed coordinates with classical, information-theoretic, Markov, semi-Markov, Bayesian semi-Markov, and endpoint-transport measures, together with held-participant/held-text prediction. The null findings define the boundary of the present evidence rather than implying that linguistic processing cannot affect microstate dynamics.
\end{enumerate}

Together, these contributions address a concrete failure mode in short-event transition analysis while making the scope of the method explicit. METER calibrates deviations after a microstate sequence has been defined. It does not make that sequence invariant to the state representation or processing stream. The reported calibration and sensitivity results therefore concern the spatially selected $K=3$ representation.

%% file: Sections/03_background.tex
\section{Methods}
\label{sec:methods}

\subsection{Theoretical motivation and relation to complementary methods}
\label{sec:theory}

Microstates are commonly obtained by clustering scalp topographies at global-field-power (GFP) peaks, back-fitting the maps to continuous EEG, and smoothing brief label fragments \citep{r1,r4,r6,r7,r43}. Within short events, the numbers of jumps and visits to each source state vary sharply, so an empirical transition table has event-specific precision. METER makes this finite-event uncertainty part of the measurement.

A second issue is interpretation. Let $J_{ij}=\Pr(X_t=i,X_{t+1}=j)$ denote the joint first-order transition flow, $s_i=\sum_jJ_{ij}$ its source-state marginal, and $A_{ij}=\Pr(X_{t+1}=j\mid X_t=i)=J_{ij}/s_i$ its conditional destination law whenever $s_i>0$. The factorization $J_{ij}=s_iA_{ij}$ shows that a joint-flow distance can change when prevalence changes even if next-state preference does not. We therefore report METER-Occ for participant-relative occupancy and METER-Jump for conditional destination choice.

Throughout, divergence is measured with the Jensen--Shannon (JS) divergence
\begin{equation}
\operatorname{JS}(p,q)=\tfrac12\operatorname{KL}(p\Vert m)+\tfrac12\operatorname{KL}(q\Vert m),\qquad m=\tfrac12(p+q),
\label{eq:js-definition}
\end{equation}
where $\operatorname{KL}(p\Vert q)=\sum_i p_i\log(p_i/q_i)$. Natural logarithms are used, so JS values are reported in nats.

Hidden Markov and semi-Markov models estimate latent dynamics and duration laws, entropy and syntax measures describe sequence organization, endpoint transport captures marginal redistribution, and conventional summaries characterize observed states. METER begins with an already labeled event and asks a different question: whether occupancy or first-order destination choices depart from a fixed participant-specific reference beyond the variation expected among events of comparable duration. The comparison experiment therefore tests empirical non-equivalence among estimands, not whether one method is universally superior.

\noindent\textbf{Proposition 1 (Target-order invariance of the evaluated endpoint bridge).}
For the endpoint transport formulation evaluated here, permutations that preserve the target endpoint marginals leave the objective unchanged. Fix the reference transition measure $Q$ and define
\begin{equation}
\mathcal{C}_Q(a,b)=\min_{\Gamma\in\Pi(a,b)}D(\Gamma\Vert Q),
\label{eq:endpoint-bridge}
\end{equation}
where $a$ and $b$ are endpoint marginals and $\Pi(a,b)$ is the set of couplings with those marginals. Any permutation of the target sequence that preserves $(a,b)$ changes neither $Q$ nor $\Pi(a,b)$; Eq.~\eqref{eq:endpoint-bridge} is therefore optimized over the same set with the same arguments and has the same value. The evaluated endpoint bridge is consequently informative about marginal redistribution, but it cannot identify altered within-event conditional jump structure. The proposition applies to the endpoint formulation evaluated here, not to every possible sequence-aware transport construction.

%% file: Sections/04_methodology.tex
\subsection{METER framework}
\label{sec:methodology}

\subsubsection{Overview}
\textit{METER} is a participant-relative event-analysis framework that separates changes in state prevalence from changes in conditional next-state choice while accounting for the finite transition evidence available within each event. Given a predefined discrete state sequence, it reserves a target-disjoint participant-specific reference bank, estimates the relevant reference-relative deviation, and calibrates that deviation against participant events matched on opportunity and design variables defined independently of the transition outcome. The general construction is defined here; the ZuCo-specific choices are consolidated in Table~\ref{tab:study-instantiation}.

\begin{figure*}[!htbp]
\centering
\fullwidthgraphic{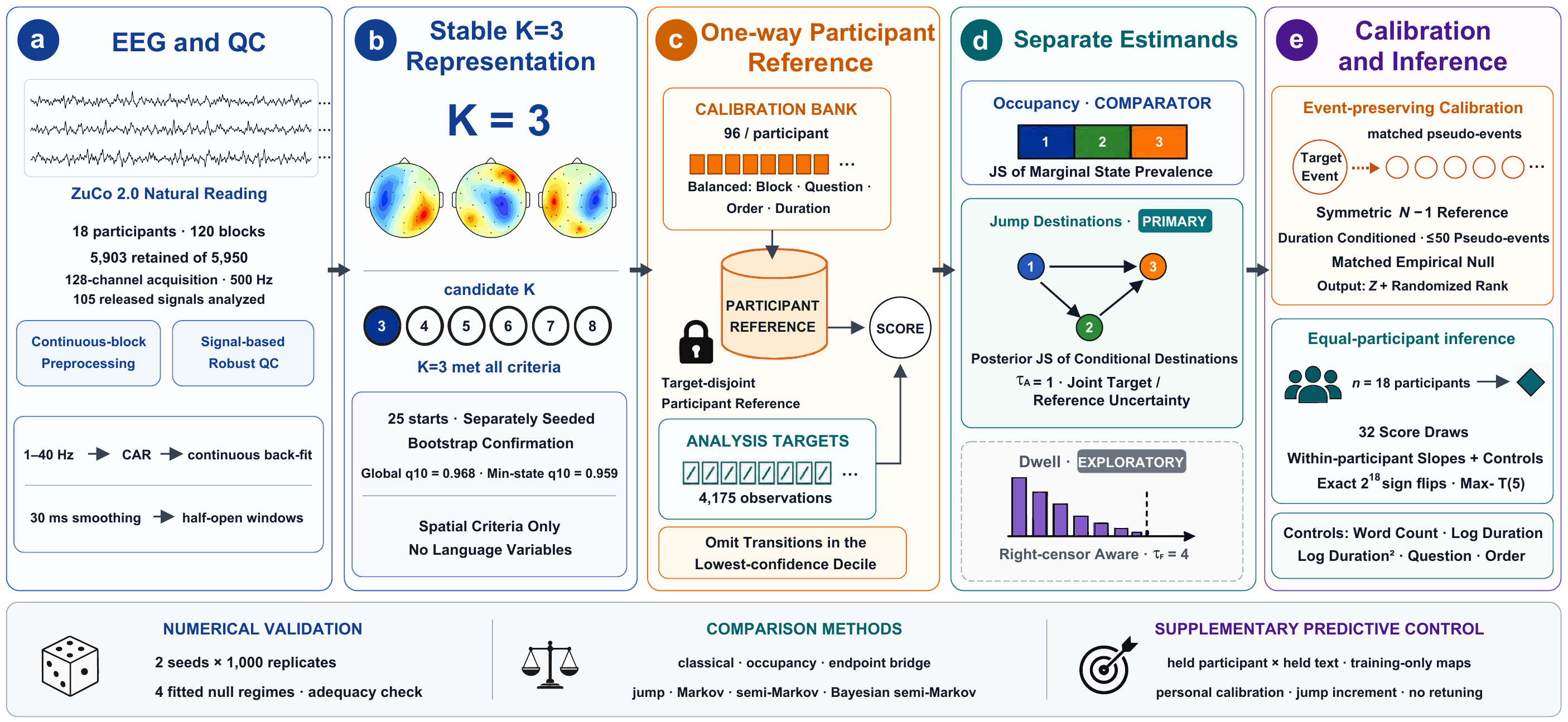}
\caption{Study-level implementation of METER. (a--b) Continuous EEG is mapped to a stable three-state representation. (c) A balanced calibration bank forms a fixed target-disjoint participant-specific reference bank. (d) METER separates occupancy, conditional jump destinations, and dwell, then calibrates each event against matched participant events before group inference (e). The lower panel summarizes validation, comparison methods, and the supplementary predictive control}
\label{fig:methodology}
\end{figure*}

Figure~\ref{fig:methodology} summarizes this analysis path. Panels a and b establish common state coordinates without linguistic outcomes. Panel c reserves a balanced bank that remains separate from every analysis target. Panel d distinguishes prevalence from conditional destination choice, with dwell treated exploratorily, and panel e carries matched event scores into equal-participant inference. Estimator and calibration choices were determined from spatial diagnostics and simulation analyses in which the linguistic variables were not used.

\FloatBarrier

For an individual event, METER-Jump asks whether its next-state choices depart from the participant-specific reference more than is typical for a comparable event. METER-Occ asks the analogous question about state prevalence. Posterior integration describes uncertainty in finite transition tables; matched-event calibration determines how unusual the resulting divergence is. Estimator ablations evaluate the distinct contributions of regularization and matching. Detailed comparator specifications are collected in Supplementary Material rather than repeated as a protocol table in the main text.

The computational sequence is
\begin{equation}
\begin{aligned}
\text{EEG}&\rightarrow K\text{ states}
\rightarrow \text{participant/target-disjoint reference}\\
&\rightarrow (R_{\mathrm{occ}}, R_{\mathrm{jump}})
\rightarrow \text{event calibration}.
\end{aligned}
\label{eq:pipeline}
\end{equation}
The primary hypothesis concerns jump-destination reconfiguration. Occupancy acts as the necessary comparator: if only METER-Occ changes, the evidence supports altered state prevalence rather than altered conditional switching. Dwell reconfiguration extends the same reference-relative logic to episode duration but remains exploratory because its power was weaker and its sensitivity to map uncertainty greater. Keeping the three coordinates separate preserves those interpretations.

\subsubsection{Microstate representation}

Let $Y_t\in\mathbb{R}^{C}$ be the $C$-channel scalp topography at time $t$. Polarity-invariant modified k-means \citep{r7} estimates maps $m_1,\ldots,m_K$ from GFP peaks by maximizing squared spatial correlation, with GFP-squared observation weights. Each continuous sample is assigned
\begin{equation}
X_t=\arg\max_{k\in\{1,\ldots,K\}}
\operatorname{corr}^{2}(Y_t,m_k)
\label{eq:backfit}
\end{equation}
so polarity does not alter the label. Assignment confidence is the margin between the largest and second-largest squared correlations. Before events are sliced, a deterministic minimum-duration smoother reassigns each sub-threshold run to the adjacent state with the strongest summed topographic evidence and merges newly contiguous runs. Temporal smoothing therefore defines which label changes count as jumps rather than serving only as denoising. METER-Occ is defined for $K\geq2$, whereas nontrivial METER-Jump reconfiguration requires $K\geq3$ because a source state needs at least two admissible destinations. Representation selection, smoothing, tie-breaking, and boundary handling for this study are specified in Section~\ref{sec:experimental} and Supplementary Material.

\subsubsection{Event-level estimands}

\begin{figure*}[!t]
\centering
\fullwidthgraphic{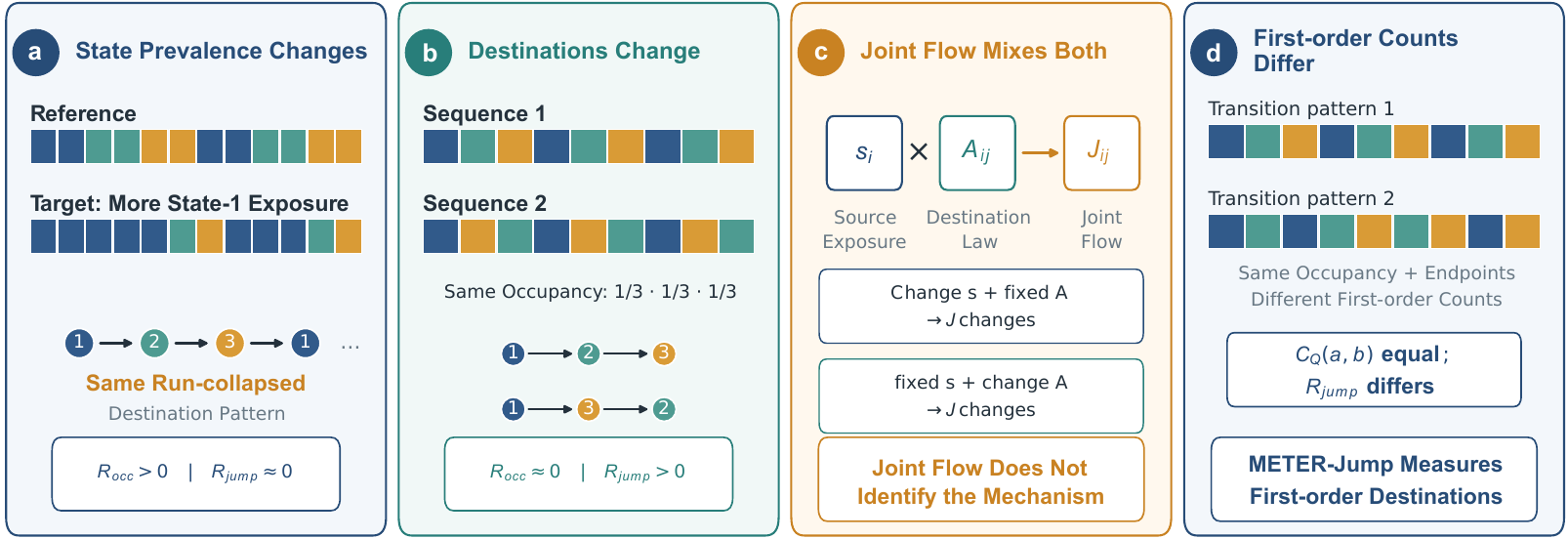}
\caption{Marginal occupancy and first-order conditional jump destinations are distinct properties of a microstate sequence. (a) State prevalence can change while the run-collapsed destination pattern remains similar. (b) Equal occupancy can accompany different destination laws. (c) The standard joint-flow decomposition $J_{ij}=s_iA_{ij}$ shows why a joint distance can change through either source-state visitation or destination preference. (d) Two sequences have equal occupancy and equal source/destination endpoint marginals but different first-order transition counts; the evaluated endpoint bridge is unchanged, whereas METER-Jump can differ. METER-Jump does not distinguish rearrangements that preserve the complete first-order edge-count table}
\label{fig:estimand-distinction}
\end{figure*}

Let $X_t\in\{1,\ldots,K\}$ denote the smoothed label in a half-open event window $[t_s,t_e)$. Marginal prevalence, conditional switching, and episode duration are represented separately. Figure~\ref{fig:estimand-distinction} shows why occupancy, conditional destinations, joint flow, and endpoint transport are not interchangeable summaries.

\paragraph{Conceptual separation of state prevalence and destination choice}

The factorization defined in Section~\ref{sec:theory} shows that $J$ can change when $s$ changes even if $A$ is identical. After consecutive self-persistence has been collapsed, the analogous source distribution is the embedded jump-chain visitation distribution $\nu$. Sample occupancy instead satisfies $\pi_i\propto\nu_i\,\mathbb{E}(D_i)$ for a stationary semi-Markov process. Occupancy and conditional jump destination are therefore separated estimands, not independent generative factors. A scalar distance between joint flows cannot, without further decomposition, identify altered conditional switching.

\paragraph{METER-Jump: jump-destination reconfiguration}

Consecutive equal labels are collapsed before jumps are counted. A jump belongs to an event only when both source and destination samples lie inside it; a state active at event onset contributes occupancy but cannot create an artificial onset jump. Let $m_{ij}$ and $n_{ij}$ be reference and target counts from source $i$ to an admissible off-diagonal destination $j$. The implemented categorical hierarchy uses Dirichlet--multinomial smoothing to avoid the severe boundary behavior of sparse plug-in probability estimates \citep{r36,r37}:
\begin{align}
Q_i\mid\mathbf m_i&\sim\operatorname{Dirichlet}(\mathbf m_i+\tfrac12\mathbf 1),\label{eq:ref-dirichlet}\\
P_i\mid\mathbf n_i,Q_i&\sim\operatorname{Dirichlet}(\mathbf n_i+\tau_A Q_i),\qquad \tau_A>0.
\label{eq:target-dirichlet}
\end{align}
For $M_i=\sum_jm_{ij}$, the fixed reference source-state weight is $w_i=M_i/\sum_kM_k$. The primary METER-Jump divergence is
\begin{equation}
R_{\mathrm{jump}}=\sum_{i=1}^{K} w_i
\mathbb{E}_{P_i,Q_i}\!\left[\operatorname{JS}(P_i,Q_i)\right].
\label{eq:jump}
\end{equation}
The expectation in \eqref{eq:jump} covers uncertainty in both finite target and finite reference tables within the fixed state representation. Each source row lies on a $K-1$ category off-diagonal destination simplex. When a target contains no outgoing jump from state $i$, \eqref{eq:target-dirichlet} remains proper and expresses uncertainty around $Q_i$; the state is neither removed nor renormalized away. The concentration and Monte Carlo resolution are implementation choices that require calibration; the values used here appear in Table~\ref{tab:study-instantiation}. Posterior draws integrate target/reference-table uncertainty at the event level; group inference is performed separately using the participant-level model in Section~\ref{sec:experimental}.

The fixed weights give greater influence to source states with more reference transitions. A change confined to a rare state can therefore contribute little to the aggregate score. Equal-state weighting represents a different estimand suited to hypotheses centered on rare states and should be validated before use.

An event receives a weak-transition-information flag when at least one source state contains too few retained outgoing jumps to represent its admissible destinations empirically. The score remains defined for an unobserved target row, but its preference is then largely prior-determined. The study-specific flag threshold and accompanying diagnostics appear in Table~\ref{tab:study-instantiation}.

\paragraph{METER-Occ: occupancy comparator}

For state $i$, define $N_i=\sum_{t=t_s}^{t_e-1}\mathbf{1}(X_t=i)$ and target occupancy $\hat p_i=N_i/\sum_jN_j$. If $M_i^\pi$ is the reference sample count, the Jeffreys-smoothed reference estimate is \citep{r36}
\begin{equation}
q_i^\pi=\frac{M_i^\pi+1/2}{\sum_k(M_k^\pi+1/2)}
\label{eq:occupancy-reference}
\end{equation}
and the occupancy comparator is
\begin{equation}
R_{\mathrm{occ}}=\operatorname{JS}(\hat p,q^\pi).
\label{eq:occupancy}
\end{equation}
For each target, $q^\pi$ uses the same target-specific 95-event subreference as METER-Jump and receives the same matched-event transformation and symmetric $N-1$ calibration. Unlike METER-Jump, its raw estimator uses empirical target occupancy and a smoothed reference without joint target/reference posterior integration. METER-Occ measures marginal state prevalence and makes no claim about transition order.

\paragraph{Exploratory dwell reconfiguration}

Only episodes beginning inside the event enter the exploratory dwell estimand. If an episode beginning at $t_m$ continues beyond event offset, its observed duration is
\begin{equation}
D_m^{\mathrm{obs}}=\min(D_m,t_e-t_m)
\end{equation}
and it contributes the right-censoring statement $D_m>t_e-t_m$. A binned survival likelihood with compatible-bin data augmentation accounts for censoring, and reference source-state-weighted JS compares target and reference duration distributions. The event null uses the same matching and symmetric $N-1$ construction as the jump score. This coordinate shows how the framework extends to duration, but it remains exploratory and is not part of the primary claim; the full construction is confined to Supplementary Material.

\subsubsection{Participant/target-disjoint reference bank}

For each participant, $N$ events form a fixed reference bank balanced over prespecified design variables. Each analysis target is scored against a target-specific $N-1$ subreference created by an outcome-independent exclusion rule and never contributes to the bank. A target therefore cannot influence the reference against which it is evaluated. ``Target-disjoint'' describes the absence of event overlap or feedback; bank and target events can still be statistically dependent because they arise from the same participant and recording context. Other studies can balance on trial type, session, run, stimulation condition, or recording quality. The ZuCo allocation variables and $N$ are given in Table~\ref{tab:study-instantiation}.

\subsubsection{Boundary-confidence filtering}

For sample $t$, let $\Delta_t=r_{t,(1)}^2-r_{t,(2)}^2$ be the upstream, pre-smoothing margin between the largest and second-largest squared map correlations. The 30-ms smoother is then applied. If the final smoothed sequence contains a jump between samples $t-1$ and $t$, its boundary confidence is
\begin{equation}
c_t=\min(\Delta_{t-1},\Delta_t).
\label{eq:transition-confidence}
\end{equation}
Thus both boundary samples must have unambiguous upstream topographic assignments. This margin is not a posterior probability for the final smoothed state. A participant-specific quantile of $c_t$ over calibration-bank jumps defines the threshold; jumps at or below it are omitted from bank and target counts. If all margins are tied, all jumps remain. The selected quantile and its empirical checks are reported in Table~\ref{tab:study-instantiation} and Supplementary Material.

The confidence threshold is estimated once from the full calibration bank and is not re-estimated during pseudo-target leave-one-out scoring. A pseudo-target therefore contributes to the shared participant threshold whereas a real target does not. Participant-fitted simulations preserve this dependence.

\subsubsection{Symmetric matched-event calibration}

Even when the underlying dynamics do not change, raw divergence tends to vary with the amount of information in an event. We estimate this finite-event background separately for each participant. Every calibration event takes a turn as a pseudo-target and is compared with the other $N-1$ calibration sequences. A real target is also compared with $N-1$ calibration sequences after one deterministic, outcome-independent exclusion. Real and pseudo-targets therefore use the same reference-bank size, although their compositions need not be identical. The reproducible hash and tie rules are given in Supplementary Material.

Matching first restricts pseudo-events by an opportunity variable defined independently of the transition outcome and may then rank them using additional design covariates. Realized jump count or source-row evidence should not define opportunity because either may change under the alternative. The ZuCo implementation uses duration quartiles and the context variables summarized in Table~\ref{tab:study-instantiation}; the complete distance, scaling, and tie rules are given in Supplementary Material.

For observed divergence $R$, null mean $\mu_0$, and null standard deviation $\sigma_0$, the reported coordinate is
\begin{equation}
Z=\frac{R-\mu_0}{\sigma_0}.
\label{eq:zscore}
\end{equation}
For $B$ matched pseudo-event scores $R_b^0$, define $N_{>}=\sum_b\mathbf{1}(R_b^0>R)$ and $N_{=}=\sum_b\mathbf{1}(R_b^0=R)$. The implemented finite-sample randomized right-tail rank is
\begin{equation}
p_{\mathrm{raw}}=\frac{N_{>}+U(N_{=}+1)}{B+1},\qquad U\sim\operatorname{Uniform}(0,1)
\label{eq:randomized-rank}
\end{equation}
where $U$ is obtained from a stable hash of the event identity and analysis seed. The $+1$ term randomizes the target's position among ties while keeping the result reproducible. Exact finite-sample validity requires exchangeability between the target and its matched pseudo-events together with symmetric scoring. Each pseudo-target is left out of the candidate score values used to rank it.

Duration matching only approximates exchangeability in this application. Real targets use 512 posterior draws whereas pseudo-targets use 128, and each pseudo-target contributes to the calibration bank from which the shared confidence threshold is derived. We therefore evaluate event-level Type-I behavior in participant-fitted simulations.

The randomized rank supplies the event-level rejection rule. By contrast, $Z$ in Eq.~\eqref{eq:zscore} is a matched-event standardized coordinate for effect estimation, not the randomized rank or an inverse-normal transform of $p$. With 24 matched pseudo-events per application target, the attainable rank grid has width $1/25=0.04$. Randomization can therefore change an individual boundary decision, but it does not enter $Z$ or the participant-level association analysis; deterministic bounds and aggregate sensitivity are reported in Supplementary Material, Table~S22.

Figure~\ref{fig:worked-score} consolidates the event path from confidence-filtered jumps through symmetric reference construction, posterior integration, event matching, and the distinct $Z$ and randomized-rank outputs; compact pseudocode is given in Supplementary Material, Algorithm~S1, together with tie handling and stored output fields.

\begin{figure*}[!t]
\centering
\fullwidthgraphic{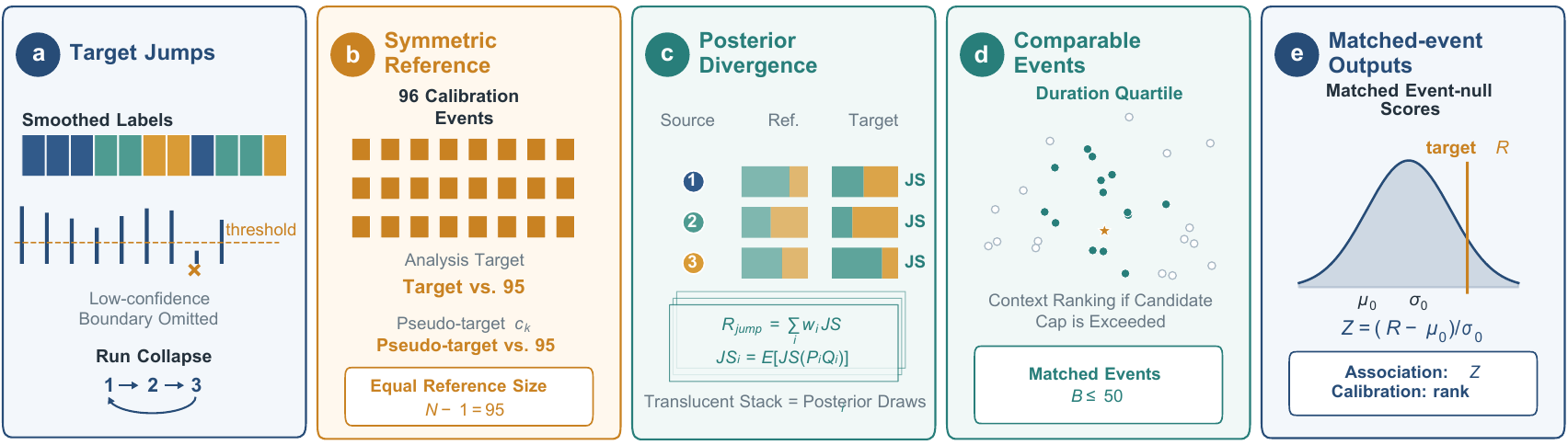}
\caption{Construction and finite-event calibration of METER-Jump. (a) Smoothed labels are collapsed into jumps after low-confidence boundaries are omitted. (b) Real and calibration pseudo-targets receive equal $N-1=95$ reference-bank size. (c) Posterior-integrated conditional-destination divergences are combined with fixed reference source-state weights. (d) Duration-first matching selects comparable pseudo-events. (e) The matched scores provide two outputs: null moments for the standardized effect coordinate $Z$, and a randomized rank $p$-value for the calibrated event-level rejection rule}
\label{fig:worked-score}
\end{figure*}
\FloatBarrier

\subsubsection{Scope and assumptions}

METER is intended for delimited events with predefined state labels, unequal transition evidence, and enough participant observations to construct a target-disjoint reference bank. Reference size, state alphabet, event definition, smoothing, and sequence reliability must be evaluated for each application. Sparse source rows, unstable maps or jump chains, unmodeled drift, and small reference banks weaken interpretation. A different state alphabet, event definition, smoother, or recording design therefore requires renewed event calibration and sequence-reliability analysis. The randomized rank is the event-level test, whereas $Z$ is the continuous coordinate used for effect estimation.

%% file: Sections/05_experimental_setup.tex
\subsection{Dataset, validation, and statistical analysis}
\label{sec:experimental}

The evaluation separates measurement validity from the naturalistic application. Spatial checks examine whether the state coordinates are reproducible, participant-fitted simulations test event-level calibration as transition information changes, ablations identify the components responsible, and the reading analysis applies METER-Jump under realistic event heterogeneity. Application associations are interpreted in light of these measurement properties.

\subsubsection{Dataset, event reconstruction, and cohort}

We analyzed the Natural Reading task of the publicly available ZuCo 2.0 corpus \citep{r20,r3}. ZuCo was acquired with a 128-channel EEG system at 500~Hz. The released cleaned task matrices used here contain 105 retained EEG signals (104 EGI-labeled electrodes and Cz); all 105 entered microstate clustering and continuous back-fitting. We did not remove a second analysis-specific channel subset. The original cohort comprised 18 healthy adult native English speakers (mean age 34 years, SD 8.3, 10 reported as female, and 16 right-handed). The corpus authors report written consent for participation and data reuse and approval by the Ethics Commission of the University of Zurich \citep{r3}. \textit{This study analyzed publicly released, de-identified data and involved no new participant recruitment.}

After three practice rows per material block were removed, 349 experimental sentences remained. Their boundaries were reconstructed from paired reading markers in the continuous EEGLAB event stream, with response markers required to fall outside the sentence window. We retained unique, ordered boundaries consistent with the block materials and treated windows outside 1--30~s as reconstruction errors before EEG quality was considered. Marker codes, indexing conversion, and validation checks are documented in Supplementary Material.

Of the 126 available block files from 18 participants, 121 passed structural reconstruction and event--material alignment checks. One additional block failed the subsequent block-level EEG-quality minimum, leaving 120 blocks for analysis. Duration and sentence-level EEG-quality criteria retained 5,903 of 5,950 eligible observations (99.21\%). Reserving 96 sentences per participant for calibration left 4,175 observations for the application and method comparisons. File-prefix handling and all six excluded-block reasons are listed in Supplementary Material.

\subsubsection{Dataset preprocessing and quality control}

\textbf{Processing supplied with the dataset:} ZuCo provides both raw EEG and recordings cleaned with Automagic 1.4.6. We used the cleaned files, in which bad channels had been interpolated and MARA had identified and removed independent components associated with ocular and muscle artifacts. The original acquisition used a 0.1--100~Hz passband and a Cz reference \citep{r3}. We retained the corpus decisions rather than rerunning ICA or relabeling its components. This inheritance is consequential because ocular-component removal can affect selected map number, map topography, and statistical power, even when more aggressive ICA choices give broadly similar results after ocular artifacts have been removed \citep{r45}. The present analysis therefore evaluates METER conditional on the released Automagic/MARA preprocessing. It is not a raw-data comparison of artifact-removal strategies.

\textbf{Additional analysis preprocessing:} We processed each continuous block before cutting it into sentences. The steps were a fourth-order 1--40~Hz Butterworth filter in second-order-section form, forward--backward zero-phase filtering, common-average rereferencing (CAR), continuous map back-fitting, a 30~ms minimum-duration smoother, and half-open sentence slicing. The passband removes slow drift and high-frequency muscle activity while preserving the frequencies commonly used for microstate topographies. CAR makes the spatial reference explicit; a separate sensitivity analysis examined the effect of omitting this additional rereferencing. Filtering and smoothing at the block level also prevent sentence boundaries from becoming artificial signal boundaries. No text information entered the remaining signal-quality decisions.

The released Cz signal was flat before added processing, consistent with its role as the acquisition reference. It remained part of the common 105-channel geometry: CAR transformed it with the other channels, whereas it remained zero in the no-additional-CAR sensitivity. All map correlations explicitly mean-centered the same 105-dimensional topographies. The near-equivalence of the two reference treatments is therefore specific to these released matrices; channel-level handling is detailed in Supplementary Material.

Median channel standard deviation was required to lie between 0.05 and 200~$\mu$V. A sentence was excluded when its 99.9th absolute-amplitude percentile exceeded 100~$\mu$V or more than 1\% of samples crossed 100~$\mu$V in any channel. These thresholds screen gross residual artifacts in the released cleaned recordings and were defined independently of the linguistic application. We evaluated a 2--20-Hz passband and omission of the additional CAR as preprocessing sensitivities. Each alternative repeated back-fitting, event reconstruction, calibration-bank construction, and METER scoring without changing the estimator or inferential rule.

Temporal sensitivity analyses used 20- and 40-ms minimum-duration rules alongside the primary 30-ms definition. Because smoothing follows map estimation, all three streams used the same group maps. Each alternative repeated continuous preprocessing, back-fitting, event extraction, calibration, and scoring for all participants while leaving the estimator and inferential hierarchy unchanged. We compared label and jump agreement, METER-Jump coordinates, calibration-bank pseudo-event rejection rates, and controlled application coefficients. Max-$T$ correction was retained within each outcome and stream, but no correction was made across processing variants. These analyses assess sensitivity rather than provide additional confirmatory tests.

\subsubsection{Study-specific METER configuration}

Table~\ref{tab:study-instantiation} separates the general METER construction from the choices evaluated in this study. These values define the study configuration rather than universal defaults.

\begin{table*}[!htbp]
\caption{Study-specific configuration of the general METER framework}
\label{tab:study-instantiation}
\centering
\small
\setlength{\tabcolsep}{4pt}
\begin{tabularx}{\textwidth}{>{\raggedright\arraybackslash}p{0.20\textwidth}>{\raggedright\arraybackslash}p{0.27\textwidth}>{\raggedright\arraybackslash}X}
\toprule
\textbf{Component} & \textbf{Study choice} & \textbf{Scientific role} \\
\midrule
State representation & $K=3$; 25 starts; 30-ms primary smoothing & Reproducible task-derived coordinates; 20- and 40-ms streams test temporal sensitivity. \\
Reference bank & 96 events per participant, jointly balanced over block, question status, acquisition order, and duration & Defines a participant-specific, target-disjoint baseline. \\
Jump estimator & $\tau_A=1$; 512 target draws; 128 pseudo-event draws & Integrates finite target/reference row uncertainty; concentration evaluated by simulation. \\
Confidence rule & Participant-bank 10th percentile of two-endpoint assignment margin & Omits the least certain jump boundaries while holding the rule fixed for targets. \\
Event calibration & Symmetric 95-event subreference; duration quartile; at most 50 pseudo-events & Gives real and pseudo-targets equal reference size and conditions on opportunity to observe transitions. \\
Weak-information flag & Fewer than two retained outgoing jumps in any source row & Identifies rows unable to represent both off-diagonal destinations empirically for $K=3$. \\
Group estimation & 32 score draws; 18 equal-weight participant slopes & Propagates conditional score variation into point estimation without treating events as independent replicates. \\
Multiplicity & Exact $2^{18}$ sign flips; max-$T$(5) within each outcome family & Controls the five linguistic associations separately for jump, occupancy, and exploratory dwell. \\
\bottomrule
\end{tabularx}
\end{table*}

\subsubsection{Representation selection and assignment uncertainty}

Map estimation used 25 starts and at most 1,000 iterations at participant and group levels. Each event contributed at most 250 GFP peaks separated by two samples, limiting domination by long events and duplicate adjacent topographies. Participant-level fits used GFP-squared weights; the group stage weighted participant maps equally. The 30-ms smoother was chosen as a conservative anti-fragmentation rule below the usual 60--150-ms duration range \citep{r1}, not as a physiological minimum. We compared $K\in\{3,\ldots,8\}$ using spatial reproducibility. Acceptance required a global 10th-percentile map correspondence of at least 0.80, a median of at least 0.90 for every state, and a state-wise 10th percentile of at least 0.70. These thresholds were based only on spatial reproducibility; downstream detection and linguistic associations did not enter representation selection. A 200-bootstrap analysis compared candidate representations, followed by a separately seeded 500-bootstrap evaluation of the selected candidate. Participant-held-out peak GEV quantified the tradeoff between reproducibility and explained variance. Supplementary Material, Tables~S1--S3 and Fig.~S1 specify these analyses and the template and smoothing comparisons.

We also compared the selected $K=3$ maps descriptively with the four-class EEG Meta-Microstates distributed with MICROSTATELAB \citep{r42,r43}. Because the resources used different montages, both were interpolated to a common scalp grid before polarity-invariant spatial correlation. These values describe correspondence rather than formal TANOVA or direct template assignment; the templates did not influence $K$, map labels, or event sequences. Supplementary Material, Table~S2 reports the full $3\times4$ matrix.

To examine uncertainty at the assignment level, we fitted another 500 bootstrap map sets and applied them to 5,400 held-out GFP peaks. This bank preserves each bootstrap map set's joint assignment and confidence pattern, and simulated labels then pass through the same 30-ms smoother as the data.

We also evaluated continuous samples to determine whether peak stability represented the whole signal. From two acquisition-spaced valid blocks per participant, we sampled topographies across participant-normalized GFP tertiles and retained one central segment from each block for temporal comparison. Five hundred participant bootstraps formed polarity-aligned, equal-weight consensus maps using the selected state identities. An approximation check compared 100 consensus maps with complete group refits. We then compared labels before and after 30-ms smoothing and measured edit distance between the run-collapsed jump chains. This analysis captures variation in participant composition and GFP level, but not within-participant peak resampling or a complete sensor-level error model; sampling and alignment details appear in Supplementary Material.

\subsubsection{Simulation and ablation design}

The simulation grid varied $\tau_A\in\{1,4,16\}$, fixed versus state-adaptive shrinkage, three confidence policies, and 96--144 reference events. Candidate configurations were evaluated primarily for Type-I calibration, with sensitivity, interval behavior, bias, and complexity as secondary criteria. Four null regimes represented participant-fitted dynamics, unequal occupancy, rare transitions, and GFP-peak-derived map perturbation. The unequal-occupancy null altered dwell while preserving destination probabilities; the rare-transition null suppressed transitions into one state; and the map null reassigned labels using bootstrap-derived maps. Each mechanism was applied to both bank and target events. A circulation alternative redirected a fraction $\lambda$ of feasible stationary flux, changing destinations while preserving embedded-chain visitation. Two separately seeded 1,000-replicate streams evaluated the chosen configuration; a separate six-point effect curve used 500 replicates per point under fixed and perturbed maps. Supplementary Material reports the configuration grid and complete generators.

Simulated transition matrices, dwell means, sentence lengths, and matching covariates came from the retained participant sequences. Additional 500-replicate checks introduced within-event transition drift or destination-dependent dwell while applying the same mechanism to banks and targets. Every target used the same 512-draw score and duration-conditioned calibration as the data.

\subsubsection{Estimator and reference-design ablations}

Paired component simulations compared plug-in JS, shrinkage-only JS, posterior-integrated JS, and METER-Jump on identical banks and targets. Global 95th-percentile cutoffs for the first three variants were learned from participant-fitted nulls and then transferred unchanged to the heterogeneous regimes; METER-Jump used its randomized matched-event rank. A separate reference experiment compared the fixed bank with LOO and balanced two-fold formulations under a shared shift relative to the fixed baseline. The designs, replicate splits, and empty-row convention appear in Supplementary Material.

First- and second-order jump models were also compared by balanced two-fold prediction on 350,399 held-out jump triples. A 200-replicate participant-fitted study anchored the observed gain against null and simulated second-order effects; its simulation-based relevance threshold is given in Supplementary Material.

\subsubsection{Numerical validation experiments}

Each main null regime contained 1,000 participant-cycled replicates per seed. Calibration required a 95\% Wilson interval containing 0.05 and excluding 0.10; participant and duration strata were also examined. Common simulated targets evaluated 32--144 reference events for error, detection, bias, interval coverage, and sparse-row behavior. The null-centered event-score interval subtracts the matched pseudo-event mean from the 2.5th and 97.5th percentiles of the posterior divergence draws; its formal definition and scope appear in Supplementary Material.

Sixteen thousand simulated group fits with $n=18$ compared an artificial single-score condition with added score uncertainty against integration over 32 score draws across null and nonzero effects. Separate analyses examined 16--128 retained draws, 8--50 matching neighbors, duration-stratified versus global matching, and 20 independently balanced reference banks. These studies assessed group recovery, numerical convergence, matching behavior, and reference-composition sensitivity without using linguistic outcomes (Supplementary Material).

\subsubsection{Naturalistic EEG application and statistical analysis}

The naturalistic application tests whether METER coordinates covary with five sentence properties spanning readability, syntactic complexity, and lexical predictability. Flesch--Kincaid grade indexed surface readability \citep{r10,r24}, parse-tree depth and mean dependency length indexed syntactic complexity \citep{r14,r25,r28}, and mean and summed GPT-2 surprisal indexed lexical predictability \citep{r17}. Surprisal was measured in bits, dependency features came from spaCy, and readability from textstat. Because GPT-2 received no preceding-sentence context, the first word was assigned zero surprisal and word count entered every model as a control. Supplementary Material, Fig.~S3 shows the predictor structure; the first-word and alternative-parser analyses retained the same model and decision rule.

Group inference followed four steps: fit the controlled association separately within each participant, propagate METER-Jump score uncertainty through 32 draw-specific slopes and average them, give the resulting 18 participant slopes equal weight, and enumerate all participant-level sign flips while controlling the maximum statistic across the five predictors.

Within each participant, we standardized the predictors, neural outcomes, word count, log duration, and acquisition order. Each linguistic predictor entered a separate model; the five predictors were not mutually adjusted. A separate slope related each predictor to each neural coordinate while adjusting for word count, linear and quadratic log duration, control-question status, and acquisition order. The coefficients are therefore controlled marginal associations rather than unique effects. The intercept absorbs participant-level baseline differences associated with the common 96-event reference bank; target-specific 95-event subreferences vary only through the stable exclusion rule. Each fitted design was required to have full column rank, and its matrix condition number was stored as a collinearity diagnostic. For jump and dwell, each of the 32 retained score draws was standardized with the same event-null moments and entered the same participant-level regression. For participant $u$ and predictor $k$, we averaged the draw-specific slopes, $b_{uk}^{(d)}$, as $b_{uk}=32^{-1}\sum_{d=1}^{32}b_{uk}^{(d)}$. Within-draw variances and between-draw slope variation were retained as diagnostics. They were not combined with Rubin's rules, and sign-flip inference was not repeated for every draw. Occupancy has one empirical score and no draw dimension. The observed equal-participant statistic is
\begin{equation}
T_k=\frac{\bar b_k}{\sqrt{\sum_u(b_{uk}-\bar b_k)^2/[n(n-1)]}}.
\label{eq:participant-t}
\end{equation}
For every sign vector $g\in\{-1,1\}^{18}$, we replace $b_{uk}$ by $g_ub_{uk}$, recompute all five statistics jointly, and record $T_{\max}^{(g)}=\max_{k=1,\ldots,5}|T_k^{(g)}|$. The adjusted two-sided value for predictor $k$ is $2^{-18}\sum_g\mathbf{1}[T_{\max}^{(g)}\geq|T_k|]$. Under sign exchangeability this enumeration is exact and follows established permutation-inference and resampling-based multiplicity principles \citep{r38,r39}. Correction is applied separately to the five jump, five occupancy, and five dwell predictors. We report the usual participant-slope $t$ interval, $\bar b_k\pm t_{.975,17}\operatorname{SE}(b_k)$, as a descriptive compatibility summary. It is not an interval obtained by inversion of the exact sign-flip test, so its zero crossing need not coincide with the exact or max-$T$ decision. A group-level null simulation preserved each participant's complete design and all 32 score draws, jointly permuted sentence rows within participant, recomputed the five integrated slopes, and applied the same $2^{18}$ max-$T$ rule in 500 replicates. Because the 349 items are treated as fixed, this analysis generalizes across participants reading these ZuCo sentences rather than across an arbitrary population of texts.

\subsubsection{Comparison methods}

We evaluated four complementary comparison families: conventional temporal summaries; reference-relative occupancy, endpoint transport, and conditional-jump distances; first-order sequence entropy; and cross-fitted Markov or semi-Markov predictive losses. All used the same saved three-state labels and 4,175 analysis observations, so differences reflect estimands rather than outcome-driven remapping. Every score entered the same participant-slope model, with Benjamini--Hochberg correction over 100 baseline--predictor tests \citep{r27}. Supplementary Material specifies each comparator's reference, regularization, censoring, and orientation.

The two plug-in first-order entropy rates are not Sample Entropy (SampEn): they describe sequence uncertainty, whereas METER-Jump describes reference-relative destination redistribution \citep{r44,r46}.

\subsubsection{Supplementary NR/TSR predictive control}

A supplementary nested held-participant/held-text NR/TSR analysis asked whether METER-Jump adds predictive information beyond text and conventional microstate features. An unlabeled bank from the held participant supplied the participant-specific reference; Supplementary Material gives the complete specification.

\subsubsection{Software and reproducibility}

The Python implementation used MNE-Python \citep{r21}, polarity-invariant routines compatible with pycrostates \citep{r22}, NumPy, SciPy, spaCy \citep{r25}, textstat \citep{r24}, GPT-2 \citep{r17}, statsmodels \citep{r26}, and scikit-learn. Configurations, provenance, environment versions, software checks, and machine-specific runtimes are reported in Supplementary Material.

%% file: Sections/06_results.tex
\section{Results}
\label{sec:results}

\subsection{Finite-event uncertainty motivates duration-conditioned calibration}

Quality control retained 5,903 of 5,950 duration-valid events (99.21\%) from all 18 participants and 120 blocks. Reserving 96 events per participant left 4,175 analysis targets with finite METER scores.

Table~\ref{tab:event-information} makes the event-information scale explicit. After boundary-confidence filtering, the median event contained 46 retained run-collapsed jumps, or 9.89 jumps/s, but the lower tail reached zero and 3.31\% of events triggered the weak-transition-information flag. Supplementary Material, Fig.~S2 shows the score distributions and participant variation. Most sentences are not extremely sparse. The challenge is that their evidence varies widely and is divided among three conditional source rows while the reference source-state distribution is participant-specific.

Confidence filtering removed a median 9.62\% of target jumps and was balanced across source states (89.74--90.30\% retained) and directed edges (8.97--10.50\% removed). Supplementary Material, Table~S18 reports participant thresholds and the full breakdown.

\begin{table}[!htbp]
\caption{Finite information in the 4,175 analysis events}
\label{tab:event-information}
\centering
\small
\begin{tabular}{lccc}
\toprule
\textbf{Quantity} & \textbf{Median} & \textbf{IQR} & \textbf{5th--95th percentile} \\
\midrule
Event duration (s) & 5.02 & 3.36--7.18 & 1.86--12.86 \\
EEG samples & 2,509 & 1,679--3,592 & 932--6,431 \\
Retained jumps & 46 & 29--68 & 14--126 \\
Retained jumps/s & 9.89 & 8.41--10.91 & 4.27--12.37 \\
Reference samples & 267,851 & 234,759--342,524 & 160,358--371,493 \\
Reference runs & 5,754 & 4,490--7,158 & 3,486--8,141 \\
\bottomrule
\end{tabular}
\end{table}

\begin{figure*}[!htbp]
\centering
\fullwidthgraphic{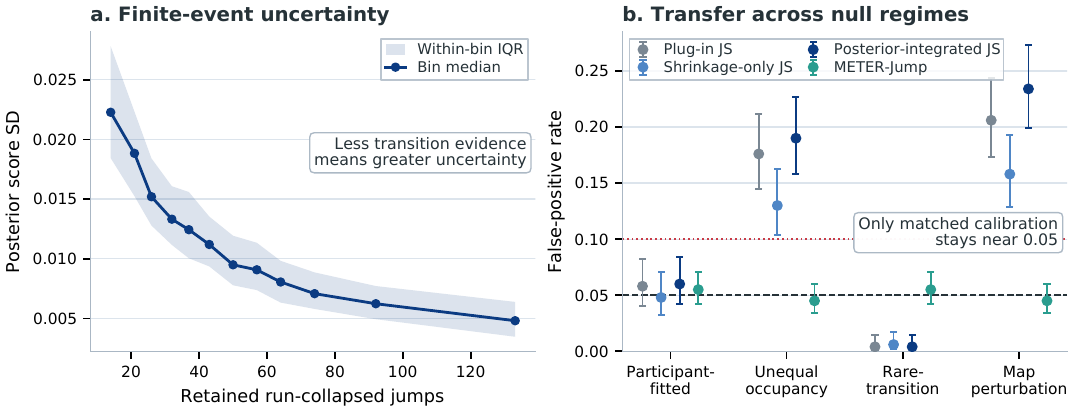}
\caption{Finite-event uncertainty and the need for duration-conditioned calibration. (a) Points and lines show the median METER-Jump posterior score standard deviation in 12 equal-frequency bins of retained run-collapsed jump count; shading spans the within-bin interquartile range. (b) Global cutoffs for plug-in JS, shrinkage-only JS, and posterior-integrated JS were learned under the participant-fitted null and held fixed when transferred to heterogeneous regimes. They do not transfer, whereas METER-Jump remains near 0.05. Intervals are 95\% Wilson intervals from paired simulations}
\label{fig:finite-information}
\end{figure*}

Figure~\ref{fig:finite-information} summarizes the central inferential problem and its solution. Posterior uncertainty is greatest when an event supplies few jumps and declines nonlinearly as transition evidence accumulates (panel a). Regularization stabilizes those sparse tables but does not by itself make a global cutoff transferable. Under unequal occupancy, false-positive rates were 0.176 for plug-in JS, 0.130 for shrinkage-only JS, and 0.190 for posterior-integrated JS; under GFP-peak-derived map perturbation they were 0.206, 0.158, and 0.234. METER-Jump kept all four rates within 0.045--0.055 (panel b; Supplementary Material, Table~S10). Across null regimes, correlations with duration ranged in magnitude from 0.44 to 0.78 for the globally thresholded scores and from $-0.091$ to 0.024 for METER-Jump. Matching, rather than regularization alone, substantially reduced the duration dependence that prevented global-cutoff transfer.

\subsection{Stable group maps still yield uncertain event-level labels}

\begin{figure*}[!htbp]
\centering
\fullwidthgraphic{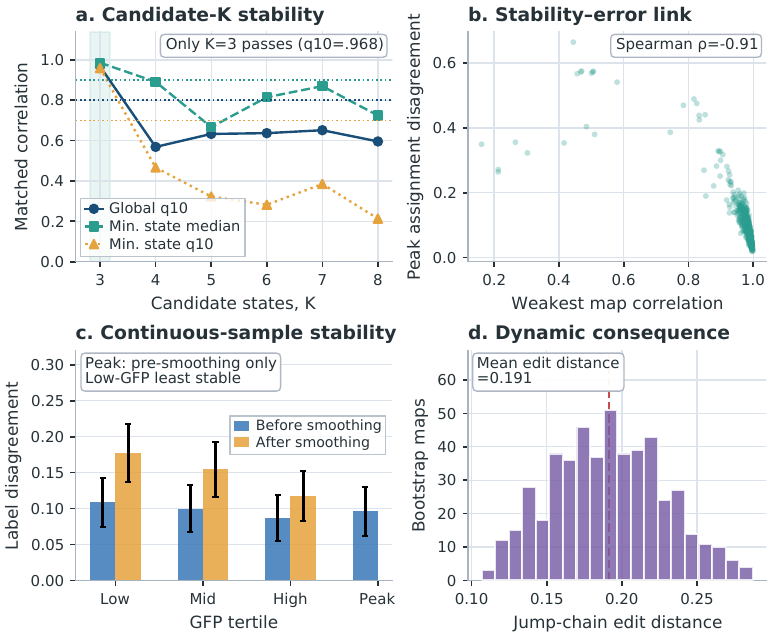}
\caption{Spatial reproducibility and assignment uncertainty. (a) Only $K=3$ met all spatial-stability criteria in the separately seeded evaluation analysis. (b) Peak assignment disagreement increases as the weakest matched-map correlation declines. (c) Bars show mean continuous-sample disagreement and whiskers show $\pm1$ SD across bootstrap map sets; the isolated peak analysis has no post-smoothing counterpart. (d) The histogram shows bootstrap-map jump-chain edit distances and the dashed line marks their mean. Reproducible group maps therefore do not imply error-free dynamic labels}
\label{fig:representation}
\end{figure*}

The three-state representation explained mean sentence-level GEV of 0.5362 and was the only candidate to meet every spatial-reproducibility criterion in the separately seeded 500-bootstrap evaluation (global 10th percentile 0.968; least-state 10th percentile 0.959, Fig.~\ref{fig:representation}a). Although $K=4$ modestly increased held-out peak GEV, its weakest-state 10th-percentile correspondence was only 0.440 in the 200-bootstrap analysis and remained poor at 0.467 in the separately seeded evaluation (Supplementary Material, Table~S1 and Fig.~S1). The choice therefore favored repeatable coordinates over maximum explained variance. State 1 corresponded most strongly with Meta A ($|r|=0.883$), whereas States 2 and 3 both matched Meta B most strongly ($|r|=0.654$ and $0.956$, respectively). This many-to-one pattern does not support a simple one-to-one assignment to the classical A--D classes, so we retained study-specific state labels (Supplementary Material, Table~S2).

Spatial stability did not eliminate sample-level uncertainty. Independently fitted maps disagreed on 11.39\% of held-out GFP-peak labels before smoothing (Fig.~\ref{fig:representation}b). The consensus-map approximation agreed with complete group refits at a median minimum-state correlation of 0.968 (10th percentile 0.898). In the continuous-sample analysis, disagreement was greatest at low GFP and increased after smoothing; mean normalized jump-chain edit distance was 0.1913 (Supplementary Material, Table~S12). The smoother removed brief fragments but could propagate a changed assignment to neighboring samples. Reproducible group maps can therefore still yield uncertain transition counts.

\FloatBarrier

\begin{figure*}[!htbp]
\centering
\fullwidthgraphic{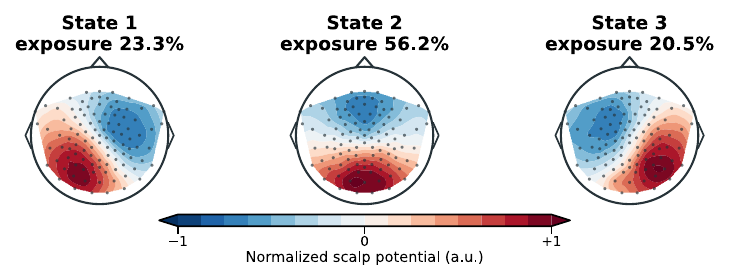}
\caption{$K=3$ group topographies. The displayed percentage is the sample-weighted prevalence of each state across retained events. A common symmetric color scale is used for all maps. Descriptive correspondences with meta-microstate templates are reported in Supplementary Material, Table~S2}
\label{fig:k3maps}
\end{figure*}

Figure~\ref{fig:k3maps} shows the spatial representation used throughout the study. The state numbers identify task-derived coordinates and do not assert one-to-one correspondence with a classical taxonomy.

\FloatBarrier

\subsection{Matched-event calibration controls false positives across unequal event information}

\begin{figure*}[!htbp]
\centering
\fullwidthgraphic{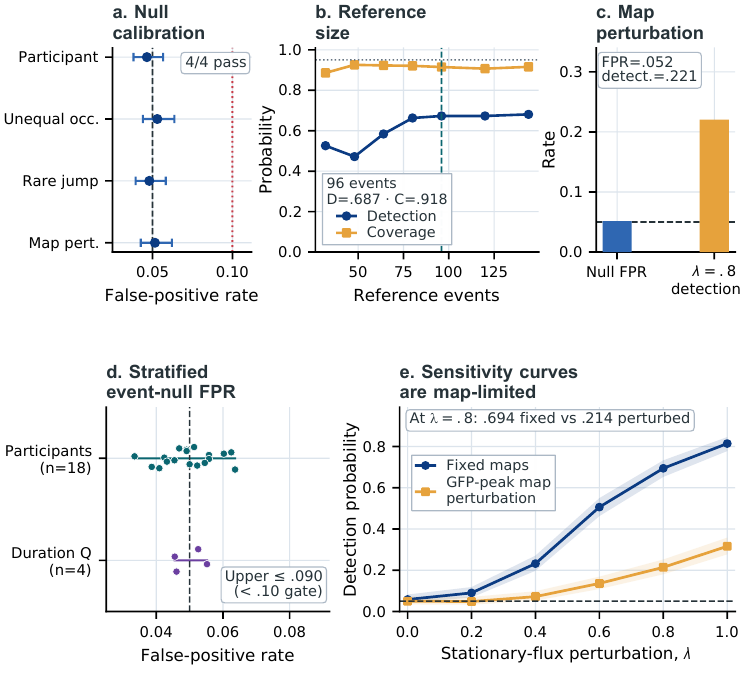}
\caption{Calibration and sensitivity. (a,d) Event-level false-positive rates with 95\% Wilson intervals for the separately seeded 96-event evaluation. (b) Common-target fixed-map detection and nominal interval coverage from 32 to 144 events; at 96 events, detection was 0.673 and coverage was 0.915, whereas the separately seeded evaluation yielded 0.687 and 0.918. The 32-event point uses global matching. (c) Map-perturbation results for the separately seeded evaluation cell. (e) A separate six-point effect-strength curve under fixed and GFP-peak-perturbed maps}
\label{fig:calibration}
\end{figure*}

The separately seeded simulation evaluation showed that the METER-Jump randomized event rank maintained the intended false-positive rate across the tested mechanisms (Supplementary Material, Table~S9; Fig.~\ref{fig:calibration}a). Across four regimes and two seeds, rates ranged from 0.044 to 0.056. Every Wilson interval included 0.05 and remained below the 0.10 inflation boundary. Deliberately misspecified time-varying and destination-dependent-dwell nulls gave rates of 0.054 and 0.060, respectively. In the separate group-level null simulation, at least one of five jump predictors crossed max-$T$ in 0.054 of replicates (95\% Wilson interval 0.037--0.077; Supplementary Material, Table~S21). These results support event-null calibration for the evaluated mechanisms rather than a distribution-free guarantee.

\subsection{Reference size and map uncertainty define sensitivity}

With 96 reference events and fixed maps, the primary $\lambda=0.8$ stationary-flux perturbation was detected in 68.70\% of the separately seeded simulations with little bias (Fig.~\ref{fig:calibration}b; Supplementary Material, Table~S13). Across participants, this perturbation produced a median largest absolute off-diagonal transition-probability change of 0.280 (IQR 0.243--0.315) and a median stationary-weighted JS change of 0.0349 nats (IQR 0.0279--0.0469; Supplementary Material, Table~S11). Nominal 95\% event-score intervals covered 91.80\% of targets in the primary 96-event evaluation and remained around 91\% at larger reference sizes (90.7\% at 120 events and 91.6\% at 144 events). The randomized event test was calibrated, whereas nominal event-score intervals showed modest residual undercoverage.

The common-target reference-size analysis located the practical plateau. Null false-positive rates remained between 0.040 and 0.064, while fixed-map detection was 0.663 at 80 events, 0.673 at 96, and 0.681 at 144 (Fig.~\ref{fig:calibration}b; Supplementary Material, Table~S13). Each point used 1,000 common targets, giving binomial Monte Carlo standard error at most 0.016. The 32-event setting invoked global matching and therefore represents a distinct evaluated setting rather than a smaller point on the same curve. Under fixed maps and this $K=3$ event distribution, 80--96 events was the lowest evaluated duration-conditioned range near the detection plateau. We therefore used 96 events in the application; linguistic outcomes played no role in that choice.

Map perturbation sharply reduced sensitivity. In the 2,000-replicate evaluation cell, detection at $\lambda=0.8$ fell to 22.05\%. The six-point curve rose monotonically from 0.058 to 0.814 under fixed maps and from 0.050 to 0.316 under the GFP-peak-derived map perturbation (Fig.~\ref{fig:calibration}c,e). This perturbation is derived from assignments at GFP peaks and is not a complete continuous-sensor uncertainty model. The continuous-sample analysis instead shows that low-GFP assignments can be less stable.

The matching sensitivity analysis behaved consistently across neighborhoods (Supplementary Material, Table~S14). Among 1,728 leave-one-out calibration pseudo-targets, duration-conditioned rejection rates were 0.0434--0.0492 across 8--50 neighbors. Because a duration quartile usually contained fewer than 50 candidates, the final matcher retained every eligible candidate; changing the block/question penalties did not alter target coordinates. Global matching was slightly more conservative (0.0411) and appreciably more duration-dependent. Across 20 balanced banks, overlapping event scores correlated 0.931--0.946 and pseudo-event rejection rates remained between 0.0446 and 0.0527. Participant-slope correlations were lower, 0.557--0.755, and the largest group-coefficient change was 0.0528; no bank changed a corrected conclusion (Supplementary Material, Table~S23). Bank composition therefore affected effect magnitude more than event rank.

\subsection{Reference construction changes the estimand}

The separate reference-design analysis addresses what counts as a departure, rather than how finite-event variation is calibrated (Supplementary Material, Fig.~S7). Fixed-bank, LOO, and balanced two-fold formulations had comparable null rates (0.0468, 0.0494, and 0.0477). Under a shared shift relative to a fixed baseline, the fixed-bank design retained the contrast, with detection probability 0.5847. Detection probabilities were 0.0319 and 0.0311 for the LOO and two-fold designs because their reference sets included the same simulated shift. This result supports the target-disjoint fixed-baseline question studied here; it does not imply that fixed banks are preferable for prediction or for every longitudinal estimand.

\subsection{Processing sensitivity reveals coordinate dependence despite stable internal rejection rates}

\begin{figure*}[!htbp]
\centering
\fullwidthgraphic{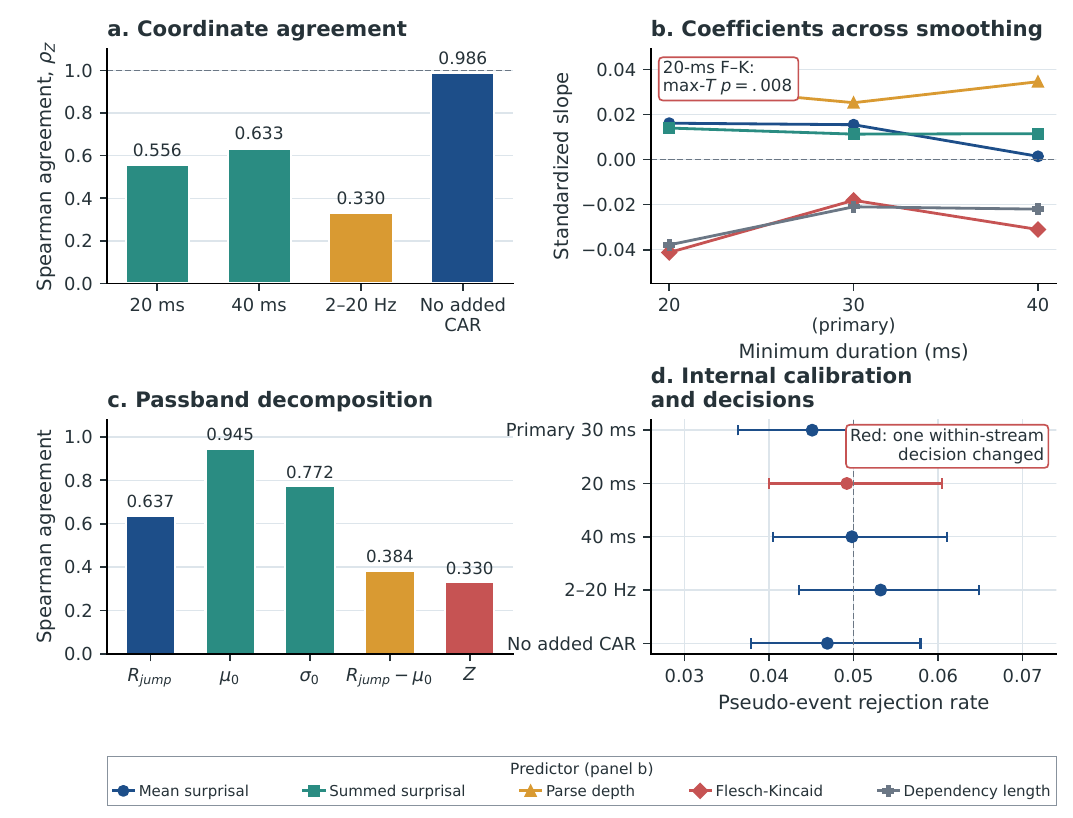}
\caption{End-to-end processing sensitivity. (a) Rank agreement of the METER-Jump $Z$ coordinate with the primary 1--40-Hz, 30-ms, CAR stream. (b) Controlled linguistic-association coefficients across the three smoothing definitions. (c) For the 2--20-Hz stream, agreement is progressively lower from raw divergence through null centering to standardized $Z$; $\mu_0$ and $\sigma_0$ are the matched-null mean and standard deviation. (d) Internal calibration-bank pseudo-event rejection rates remain near 0.05, yet the 20-ms stream changes one within-stream max-$T$ decision. Error bars in panel d are 95\% Wilson intervals for 1,728 pseudo-targets; this is an exchangeability check rather than a processing-specific participant-fitted simulation, and processing variants were not jointly multiplicity-corrected}
\label{fig:processing-sensitivity}
\end{figure*}

Processing sensitivity analyses revealed that internal pseudo-event calibration and coordinate stability are different properties (Fig.~\ref{fig:processing-sensitivity}). Under 2--20~Hz filtering, event GEV and jump counts remained strongly associated with the primary analysis ($\rho=0.975$ and 0.964), but METER-Jump $Z$ agreement over 4,167 common events was low ($\rho=0.330$). The matched-null mean was highly stable ($\rho=0.945$). The principal change arose in raw event divergence ($\rho=0.637$), and subtracting the event-specific null center reduced concordance further ($\rho=0.384$). The low $Z$ agreement cannot be attributed simply to the null-scale denominator ($\rho=0.772$). The internal pseudo-event rejection rate was 0.0532, all five coefficient signs persisted, and no corrected application decision changed (Supplementary Material, Tables~S16 and S19). Aligned map correlations were 0.9989--0.9995, showing that near-identical maps did not imply invariant event coordinates.

Omitting the added CAR was much less consequential for these already cleaned, reference-coded matrices: $Z$ agreement was $\rho=0.986$, the pseudo-event rejection rate was 0.0469, and every sign and corrected conclusion was retained. Map-stability and common-event GEV correlations exceeded 0.9999. This finding is specific to the released flat reference signal and mean-centered spatial-correlation geometry. It should not be generalized to arbitrary rereferencing of raw EEG.

Temporal regularization changed the observed jump process more visibly. Before boundary-confidence filtering, median run-collapsed jump counts were 82, 51, and 36 under the 20-, 30-, and 40-ms definitions; agreement with the primary $Z$ coordinate was $\rho=0.556$ and 0.633 at 20 and 40~ms. Internal pseudo-event rejection rates were 0.0492 and 0.0498. At 20~ms, Flesch--Kincaid grade crossed the within-stream jump threshold ($\beta=-0.0413$, t-summary 95\% interval $-0.0652$ to $-0.0173$; max-$T$ $p=0.0082$) and the exploratory dwell threshold ($p=0.0049$); at 40~ms no corrected decision changed. These sensitivity $p$-values retain five-predictor correction within each outcome stream but are not corrected across processing variants. Because each linguistic predictor was fitted separately, the Flesch--Kincaid coefficient is a controlled marginal association, not a unique readability effect; it requires evaluation in independent data.

The numerical METER-Jump coordinate therefore depends on preprocessing and state-sequence construction even when the internal pseudo-event rejection rate remains near 0.05.

\subsection{Adjacent methods measure different properties of the sequence}

Second-order context improved held-out log score by only 0.00205 nats per jump across 350,399 observations, negligible relative to both the null distribution and simulated moderate-to-strong second-order effects (Supplementary Material). We therefore retained the first-order projection for these events. Entropy-rate, Sample Entropy (SampEn), and higher-order syntax approaches remain complementary descriptions of sequence organization \citep{r44,r46}.

All comparison methods were evaluated on the same 4,175 non-calibration observations. They were empirically non-equivalent because they describe different properties of the same labeled sequence; duration and rank-correlation results are reported in Supplementary Material, Fig.~S6. The endpoint Schr\"odinger bridge was almost rank-equivalent to occupancy KL ($\rho=0.9965$ overall; median within-participant $\rho=0.9977$; Supplementary Material, Fig.~S6c). The endpoint-order proposition in Eq.~\eqref{eq:endpoint-bridge} explains this behavior: the evaluated endpoint bridge reflects marginal redistribution, whereas METER-Jump isolates conditional destination reconfiguration.

No comparator association survived its broader 100-test BH screen; the smallest $q$ was 0.7166 (Supplementary Material, Fig.~S4 and Table~S17). The sample-sequence and jump-sequence first-order entropy-rate rows were also uniformly nonsignificant ($q=0.8974$--0.9951). We validated the Barzon endpoint-transport implementation against the archived analysis: the converged solver correlated 0.99996 with the reported costs, and the inferential statistics matched the archived results. This confirms implementation fidelity but does not imply equivalence between the endpoint estimand and METER-Jump.

\FloatBarrier

\subsection{Score-draw integration supports group-level estimation}

Averaging participant slopes over score draws improved group-model recovery. Across 16,000 simulated fits with $n=18$, null rejection rates were 0.048--0.052 and null-parameter coverage was 0.948--0.952 in the balanced and participant-imbalance scenarios; coverage across their individual nonzero-effect cells ranged from 0.946 to 0.963. In an artificial single-score condition with added score uncertainty, coverage fell to 0.410 at the largest tested effect, whereas 32-draw integration yielded 0.947--0.961 coverage across nonzero effects. Increasing retained draws from 32 to 128 changed no group slope by more than 0.00254 and did not alter a max-$T$ conclusion (Supplementary Material, Tables~S15 and S21).

\subsection{Naturalistic-reading associations are null in the primary analysis but sensitive to smoothing}

\begin{figure*}[!htbp]
\centering
\fullwidthgraphic{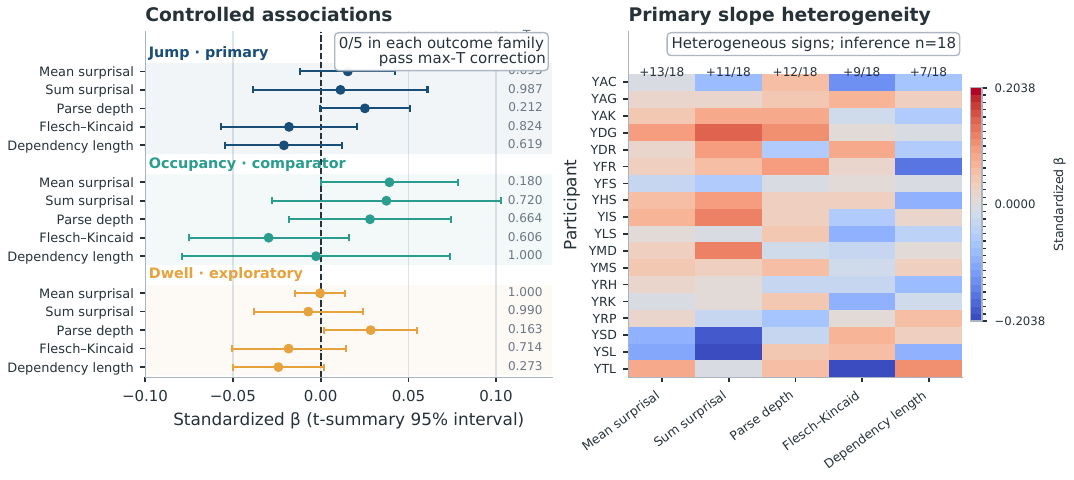}
\caption{Naturalistic application estimates and participant heterogeneity. Equal-participant standardized slopes with descriptive participant-$t$ 95\% compatibility intervals and within-outcome max-$T$ $p$-values are shown beside participant-level METER-Jump slopes. No predictor passed correction in any outcome family. The intervals summarize compatibility but are not inversions of the exact sign-flip test}
\label{fig:associations}
\end{figure*}

The naturalistic-reading application asked whether the duration-conditioned standardized $Z$ coordinate was consistently associated with sentence properties after controlling duration and the remaining design covariates. All participant designs were full rank; their participant-wise maximum condition numbers ranged from 4.75 to 6.52 (median 5.03). Participant-level, rather than event-level, replication determined inference ($n=18$; Fig.~\ref{fig:associations}). In the primary 30-ms analysis, none of the five METER-Jump associations survived max-$T$ correction. Parse-tree depth produced the largest controlled marginal estimate, $\beta=0.0252$ (t-summary 95\% interval $-0.0003$ to 0.0507; adjusted $p=0.2124$), with heterogeneous signs across participants. No METER-Occ or exploratory dwell association survived its corresponding correction. Excluding the first word from mean surprisal changed no jump coefficient by more than 0.00335 (minimum max-$T$ $p=0.2151$), and the transformer parser changed none by more than 0.00408 (minimum max-$T$ $p=0.2678$); neither changed a decision. The 20-ms definition changed one within-stream max-$T$ jump decision, whereas the 40-ms, 2--20-Hz, and no-additional-CAR analyses did not (Fig.~\ref{fig:processing-sensitivity}; Supplementary Material, Table~S16). Leave-one-participant-out estimates preserved every full-sample direction (Supplementary Material, Table~S6); full estimates appear in Table~S4.

Although the application contains 4,175 events, its population-level precision and generalizability are set by 18 individuals. The parse-depth estimate was positive but imprecise, and its descriptive interval included zero; the simulations also show that map variation limits detection. The crossed participant-and-text control gave the same practical result: adding METER-Jump changed balanced accuracy by $-0.0019$ beyond conventional microstate features and by 0.00014 beyond text plus conventional features (Supplementary Material, Table~S5 and Fig.~S5). The application is a real-data test under heterogeneous event information, rather than evidence that language cannot influence microstate dynamics.

%% file: Sections/07_conclusion.tex
\section{Discussion}
\label{sec:discussion}

METER addresses a recurring measurement problem in event-level microstate analysis: empirical transition distances change with the amount and allocation of observed information even when the underlying process does not. Across participant-fitted simulations of the selected three-state representation, matched events supplied an empirically calibrated rejection rule where global cutoffs failed to transfer. The same experiments exposed the method's boundary. Map perturbation reduced sensitivity sharply, and one naturalistic-reading conclusion changed with temporal smoothing. The resulting interpretation follows four evidential layers: \textit{map reliability, sequence reliability, event-test calibration, and downstream effect stability}. Success at one layer does not establish the next.

\subsection{Duration-conditioned calibration addresses finite-event variation separately from sequence uncertainty}

Transition probabilities are among the least reliable conventional microstate characteristics \citep{r40}. Unequal finite transition evidence is one tractable contributor. Separating occupancy from destination choice keeps prevalence and switching claims distinct, while posterior integration stabilizes sparse target and reference rows. The component experiment identifies the essential inferential step: within-participant duration conditioning greatly reduced the observed dependence on event information and restored near-nominal error across the evaluated nulls; shrinkage alone did not.

Calibration is not information recovery. Once uncertain map assignment or temporal averaging changes the observed jump sequence, downstream calibration cannot reconstruct the missing dynamics. False-positive behavior may remain nominal while sensitivity declines or the continuous coordinate moves. In the present study, the randomized event rank was calibrated across the evaluated null mechanisms, but the nominal 95\% score interval covered 91.8\%. Event-test calibration, interval calibration, and propagation of uncertainty into group effects are therefore distinct claims.

The task-derived $K=3$ maps are study-specific measurement coordinates rather than classical functional classes. They were selected for spatial reproducibility, not resemblance to the classical A--D microstate taxonomy. State 1 corresponded strongly with Meta A ($|r|=0.883$), while States 2 and 3 both matched Meta B most strongly ($|r|=0.654$ and $0.956$). This many-to-one correspondence does not form a simple three-class subset of the common four-template system. Because the comparison required interpolation between different measured montages, assigning classical A--D labels would require a direct common-montage template-assignment analysis.

\subsection{Complementary estimands answer different questions}

METER-Jump measures first-order, participant-relative destination reconfiguration. Sequences with the same first-order edge-count table receive the same jump description even if their multi-step ordering differs. Entropy-rate, Sample Entropy (SampEn), and higher-order syntax methods supply that complementary view of sequence organization \citep{r44,r46}. The sample-sequence and jump-sequence comparators used here are plug-in first-order entropy rates. Because second-order context added negligible held-out predictive information, the first-order projection was adequate for this dataset and event definition; other tasks may require a richer order.

The near-equivalence between endpoint bridge cost and occupancy divergence illustrates why this distinction matters. In the evaluated implementation, endpoint transport summarizes marginal redistribution. METER-Jump asks whether conditional destinations changed under fixed reference source-state weights, whereas Markov and semi-Markov losses quantify held-out prediction. Their different empirical behavior follows from different scientific questions; the estimand, rather than a universal performance ranking, determines the appropriate method.

The reference-design comparison has the same narrow interpretation. A predesignated target-disjoint bank retained a simulated population-wide shift relative to a fixed baseline, whereas the LOO and two-fold reference sets included the same simulated shift and absorbed it. These complete formulations differ in reference composition and calibration. Cross-fitting may remain preferable for prediction or other longitudinal estimands.

\subsection{Reference construction and preprocessing define the evaluated measurement}

Repeating the analysis across 20 balanced banks revealed two reliability levels: event-score ranks were highly stable, whereas participant slopes were more reference-sensitive. Effect-magnitude studies should therefore repeat the balanced construction and report between-bank variation. The current group model conditions on one bank and does not yet propagate this source of uncertainty formally.

The reference-size curve provides a practical guide rather than a universal sample-size rule. Detection was already close to its observed plateau by 80--96 reference events (0.663--0.673, compared with 0.681 at 144), whereas the 32-event condition required a different global matcher. For this $K=3$ event distribution, 80--96 events was therefore the lowest evaluated duration-conditioned range near the plateau. Reserving 96 events used about 30\% of each participant's eligible observations, creating a real tradeoff between reference precision and the data left for association analysis.

The matching analysis clarifies the effective calibration rule. A duration quartile usually contained fewer than 50 candidates, so all eligible candidates were retained and the block/question penalties did not alter target coordinates. In this dataset, METER therefore operated as participant-specific duration-stratified calibration, with duration defined independently of the transition outcome as the opportunity variable. Larger banks may activate the remaining context terms and would require renewed validation.

The end-to-end sensitivities separate internal calibration behavior from stability of the scientific decision. A 2--20-Hz passband changed METER-Jump coordinates substantially even though the calibration-bank pseudo-event rejection rate remained near 0.05 and all coefficient signs and corrected conclusions were retained. Omitting added CAR changed little, but this result is specific to the corpus's flat reference signal and correlation geometry. Alternative smoothers agreed only moderately with the 30-ms coordinate: 40~ms retained the primary decision, whereas 20~ms crossed one within-stream boundary. Because those $p$-values were not corrected across processing variants, the isolated result does not establish 20~ms as preferable. The internal pseudo-event analysis evaluates exchangeability within the calibration bank; it is not a processing-specific end-to-end null simulation. New applications should therefore report dynamic-feature agreement, score agreement, internal pseudo-event rejection, and application-estimate stability, and should evaluate processing-specific error and power directly when inferential guarantees are required.

METER provides a calibrated event-level rejection rule conditional on the maps, reference convention, back-fitting rule, confidence policy, and smoother that define the observed sequence, together with a matched standardized coordinate for effect estimation. The numerical METER-Jump coordinate depends on those choices rather than representing a preprocessing-invariant latent quantity. Numerical comparability must be established before coordinates from different processing streams are pooled or contrasted.

The uncertainty claims must also remain specific. The randomized matched-event test was calibrated under the evaluated null mechanisms. Nominal event-score intervals showed modest undercoverage (91.8\% rather than 95\%), and the continuous-sample analysis showed substantial low-GFP label uncertainty after smoothing. Accordingly, the study supports event-test calibration, not universally calibrated uncertainty intervals or a complete sensor-level error model.

\subsection{Naturalistic reading yields a null primary result with processing sensitivity}

Naturalistic reading exposed the finite-information problem in a scientifically meaningful setting: sentence lengths, jump counts, and linguistic properties varied together, while each sentence supplied one short state sequence. No association survived correction in the primary 30-ms analysis, and no comparator association survived its broader 100-test false-discovery-rate screen. Thus, the calibrated estimator did not convert heterogeneous event information into a positive linguistic claim. One association crossed its within-stream threshold at 20~ms, identifying a processing-dependent boundary that should be tested in independent data.

Several mechanisms may limit sensitivity. Sentence aggregation can dilute word- or fixation-locked responses; a three-map representation may merge distinct states; hard assignment discretizes spatial ambiguity; and first-order destinations omit higher-order syntax \citep{r46}. The controls in Supplementary Material sharpen this boundary. In a separate public benchmark, conventional EEG summaries yielded near-chance balanced accuracy, whereas gaze and text features carried clear task information (Supplementary Material, Table~S8). The nested crossed control likewise found no reliable METER-Jump increment beyond conventional microstate features or beyond text plus conventional features (Supplementary Material, Table~S5 and Fig.~S5). Taken together, these results are compatible with limited task information in the evaluated sentence-level EEG summaries and do not isolate METER-Jump as the source of the null result. The application therefore defines an operating boundary of the present measurement, not evidence that language has no effect on microstate dynamics.

\subsection{Practical recommendations for microstate studies}

The validation suggests the reporting practices in Table~\ref{tab:recommendations}. They are conditional on the present $K=3$ study and should be revalidated when the state alphabet, smoother, preprocessing, or event definition changes.

\begin{table}[!h]
\caption{Practical recommendations arising from the validation}
\label{tab:recommendations}
\centering
\small
\hyphenpenalty=10000
\exhyphenpenalty=10000
\begin{tabularx}{\textwidth}{>{\raggedright\arraybackslash}p{0.29\textwidth}>{\raggedright\arraybackslash}X}
\toprule
\textbf{Question} & \textbf{Recommendation} \\
\midrule
Are group maps stable? & Bootstrap map correspondence and quantify sequence uncertainty; map correlation alone does not establish reliable jumps. \\
Are events information-limited? & Report realized transition evidence and calibrate within participant using an opportunity variable defined independently of the transition outcome, such as duration. \\
How large should the reference bank be? & Examine a reference-size curve under the intended state alphabet and event distribution; in this study, 80--96 events was the lowest evaluated range near the fixed-map detection plateau. \\
Is the reference participant-specific? & Keep targets out of the fixed bank and repeat balanced-bank construction when effect magnitude matters. \\
Is switching reconfiguration claimed? & Compare against occupancy and distinguish first-order destination choice from higher-order sequence organization. \\
Are preprocessing or smoothing changed? & Revalidate jump counts, score agreement, event-null error, and application estimates. \\
Are coordinates compared across pipelines? & Treat them as different operational measurements unless numerical comparability is directly demonstrated. \\
\bottomrule
\end{tabularx}
\end{table}
\FloatBarrier

\subsection{Limitations and priorities for external validation}

The empirical evidence comes from 18 participants and one public corpus. Although event-level estimation uses 4,175 sentences, participant-level generalization is based on 18 individuals reading a fixed set of texts. The study has neither an independent replication nor a physiological condition known to alter METER-Jump, so it validates a measurement procedure rather than a biological discovery. The most informative next experiment would combine a predefined physiological contrast or repeated session with a larger cohort and independently sampled events, allowing construct validity and test--retest reliability to be measured directly.

Map assignment was the largest sensitivity limitation evaluated here. The continuous-sample analysis showed a mean smoothed jump-chain edit distance of 0.191, whereas the power experiment used a separate GFP-peak-derived perturbation bank. Future validation should carry continuous bootstrap-map reassignment and within-participant peak resampling through the complete false-positive and power analysis. Probabilistic back-fitting and hierarchical sensor-level maps could carry spatial ambiguity into the sequence model rather than thresholding it away.

Reference source-state weighting prevents target occupancy from defining the jump score but downweights reconfiguration confined to rare reference states. Equal-state weighting is a scientifically meaningful alternative that would require separate calibration. Group estimates also condition on one selected reference bank; repeated-bank analyses characterize this sensitivity but do not formally propagate it into group inference. The participant-fitted nulls represent slow session structure only indirectly, and bank and target events remain dependent because they arise from the same participant and session. Supplementary Material documents additional numerical boundaries involving rank granularity, confidence filtering, Monte Carlo resolution, matching neighborhoods, and the simulation alternative.

The reading models omit fixation count, fixation duration, saccades, and regressions; a gaze-adjusted analysis would provide an important robustness test. Because correlated linguistic predictors entered separate models, their coefficients are controlled marginal associations rather than unique effects. Finally, the 349 texts are fixed items, so inference generalizes across participants reading these sentences rather than across all English sentences.

The comparison set remains deliberately estimand-focused rather than exhaustive. We reproduced the archived endpoint-transport analysis but did not execute the external Bayesian state-dynamics package; the common-representation Bayesian semi-Markov comparator is not a reproduction of that implementation. Consequently, the study supports comparison among the implemented first-order, duration-aware, and endpoint estimands, but not superiority over every available Bayesian sequence model.

\section{Conclusions}
\label{sec:conclusion}

Short microstate events do not all contain the same transition evidence. The METER framework makes that fact part of the measurement by separating state prevalence, conditional destination choice, and exploratory dwell. Its primary METER-Jump estimator integrates sparse target and reference-table uncertainty and calibrates each event against comparable observations that cannot update its participant-specific reference bank. Across the tested heterogeneous regimes, the METER-Jump randomized rank kept false-positive rates near 0.05 where global cutoffs for plug-in, shrinkage-only, and posterior-integrated scores failed to transfer.

The validation also shows what calibration can and cannot establish. For the present three-state distribution, approximately 80--96 reference events approached the fixed-map detection plateau, but bootstrap-derived map variation caused the largest loss of sensitivity. In naturalistic reading, the primary 30-ms analysis found no corrected linguistic association, whereas a 20-ms sensitivity changed one within-stream decision. The scientific conclusion is therefore tied to the operational state sequence, not merely to the downstream test.

METER turns a raw, information-dependent transition distance into a participant-relative question: was this event more unusual than events of comparable duration from the same participant? In the evaluated three-state regimes, it provides an empirically calibrated event test together with a duration-conditioned coordinate for effect estimation. Application therefore requires four separate demonstrations: reproducible maps, reliable state sequences, calibrated event-level error, and stable downstream effects. None can substitute for another. Positive physiological validation, larger and repeated cohorts, and fuller propagation of map and reference uncertainty are the next steps toward broader use.

%% file: Sections/08_supplement.tex
\appendix
\section{Representation and preprocessing validation}

\setcounter{figure}{0}
\setcounter{table}{0}
\renewcommand{\thefigure}{S\arabic{figure}}
\renewcommand{\thetable}{S\arabic{table}}

Supplementary Material supplies implementation details and sensitivity analyses for four evidential layers: representation and preprocessing, estimator construction, group/application analysis, and operating-characteristic robustness.

\begin{center}
\captionof{table}{Selection without application outcomes and separately seeded evaluation analyses for $K=3$ and $K=4$}
\label{tab:k4-rejection}
\small
\setlength{\tabcolsep}{4pt}
\begin{tabular}{lrrrr}
\toprule
& \multicolumn{2}{c}{$\boldsymbol{K=3}$} & \multicolumn{2}{c}{$\boldsymbol{K=4}$} \\
\cmidrule(lr){2-3}\cmidrule(lr){4-5}
\textbf{Quantity} & \textbf{Select. 200} & \textbf{Evaluate 500} & \textbf{Select. 200} & \textbf{Evaluate 500} \\
\midrule
Participant-held-out peak GEV & 0.5870 & -- & 0.5965 & -- \\
Global correlation $q_{10}$ & 0.9657 & 0.9680 & 0.5302 & 0.5688 \\
Lowest-state median correlation & 0.9861 & 0.9860 & 0.8756 & 0.8914 \\
Lowest-state correlation $q_{10}$ & 0.9423 & 0.9592 & 0.4403 & 0.4666 \\
Spatial-stability criteria & Pass & Pass & Fail & Fail \\
\bottomrule
\end{tabular}
\begin{flushleft}\footnotesize Select: the 200-bootstrap analysis used to identify the candidate representation. Evaluate: a separately seeded 500-bootstrap analysis performed after the selection criteria and candidate choice were fixed. Held-out GEV was computed in the selection analysis and was not an evaluation criterion. The earlier aggregate summary averaged the minimum matched-state correlation across selection bootstraps; it is omitted because the fixed selection criteria operate on the reported global and state-specific quantiles.\end{flushleft}
\end{center}

\begin{table}[!htbp]
\caption{Polarity-invariant correspondence with four-class EEG meta-microstates}
\label{tab:meta-correspondence}
\centering
\small
\begin{tabular}{lrrrr}
\toprule
 & \multicolumn{4}{c}{\textbf{Absolute spatial correlation}} \\
\cmidrule(lr){2-5}
\textbf{Study map} & \textbf{Meta A} & \textbf{Meta B} & \textbf{Meta C} & \textbf{Meta D} \\
\midrule
State 1 & \textbf{.883} & .134 & .798 & .404 \\
State 2 & .592 & \textbf{.654} & .121 & .262 \\
State 3 & .226 & \textbf{.956} & .704 & .782 \\
\bottomrule
\end{tabular}
\begin{flushleft}\footnotesize Values were computed after cubic interpolation of the 105 study scalp channels and 71 template channels to a common 160-by-160 grid, retaining 14,436 points inside normalized radius .88. Bold marks each study map's best descriptive correspondence. Because the measured montages differ, these are descriptive interpolated correlations rather than formal TANOVA or common-montage template assignments. The meta maps were not used in representation selection or back-fitting.\end{flushleft}
\end{table}

\begin{figure}[!htbp]
\centering
\includegraphics[width=\textwidth]{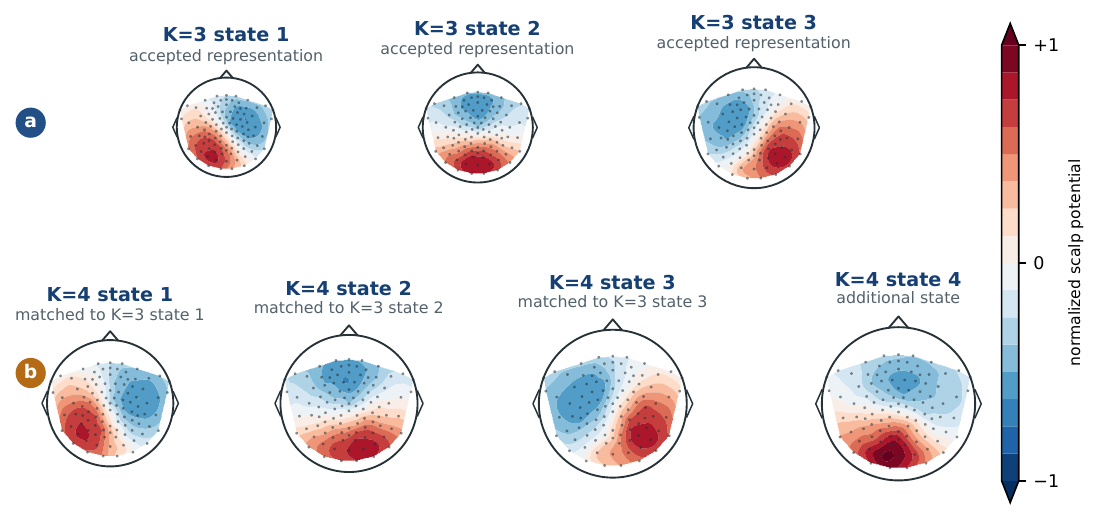}
\caption{Selection-analysis maps obtained without application outcomes for the accepted $K=3$ and rejected $K=4$ representations. The first three $K=4$ maps are polarity-aligned to their maximum-absolute-correlation $K=3$ counterparts; the fourth is the unmatched additional map. Although $K=4$ increased participant-held-out peak GEV from 0.5870 to 0.5965, its selection least-state median/10th-percentile correlations were 0.8756/0.4403, below the spatial-stability criteria. The corresponding separately seeded evaluation values were 0.8914/0.4666; $K=3$ passed both analyses (Table~\ref{tab:k4-rejection})}
\label{fig:k3-k4-maps}
\end{figure}

Table~\ref{tab:k4-rejection} separates the selection and separately seeded evaluation analyses, while Fig.~\ref{fig:k3-k4-maps} shows the maps from the selection analysis. The extra $K=4$ state increased held-out variance explanation slightly but was not reproducibly recovered across bootstrap samples. Table~\ref{tab:meta-correspondence} further shows that the three retained study maps do not support a one-to-one assignment to classical A--D classes. METER-Occ is available for $K\geq2$; nontrivial METER-Jump reconfiguration requires $K\geq3$. The false-positive and sensitivity claims reported here apply only to the selected $K=3$ representation.

\begin{center}
\captionof{table}{Temporal-smoothing feature analysis performed without application outcomes}
\label{tab:smoothing-sensitivity}
\small
\begin{tabular}{p{6cm}ccc}
\toprule
\textbf{Quantity} & \textbf{20 ms} & \textbf{30 ms} & \textbf{40 ms} \\
\midrule
Median run-collapsed jumps before confidence filtering & 82 & 51 & 36 \\
GEV Spearman $\rho$ vs. 30 ms & .9947 & 1 & .9957 \\
Jump-count Spearman $\rho$ vs. 30 ms & .9798 & 1 & .9812 \\
Jump-rate Spearman $\rho$ vs. 30 ms & .7583 & 1 & .8161 \\
Mean sample-label agreement vs. 30 ms & .9098 & 1 & .9386 \\
Median run-chain edit distance vs. 30 ms & .3673 & 0 & .2766 \\
\bottomrule
\end{tabular}
\begin{flushleft}\footnotesize All 5,903 retained sentence events from 18 participants were reprocessed with the primary group maps. Edit distance is Levenshtein distance divided by the longer run-collapsed chain. These are label-sequence feature summaries, not recalculated calibrated METER scores or application tests.\end{flushleft}
\end{center}

Table~\ref{tab:smoothing-sensitivity} shows that changing the minimum-duration rule preserved event GEV but changed the run-collapsed chain and jump-rate summaries. It therefore motivates the end-to-end score and application sensitivity analyses reported later rather than establishing interchangeability of the three smoothing definitions.

\paragraph{Block exclusions:} Six blocks were excluded. YAC block~6 (\texttt{bip}) retained only one sentence after EEG quality control and failed the block-level minimum. All 50 aligned sentences in YAK block~1 (\texttt{oip}) fell outside the configured 1--30-s duration range. EEGLAB events were not strictly ordered in YTL blocks~3 (\texttt{gip}), 4 (\texttt{oip}), 5 (\texttt{gip}), and 6 (\texttt{oip}). Prefixes are retained source-file identifiers rather than inferred quality categories. Five failures arose during structural evaluation; YAC block~6 passed reconstruction but failed the later EEG-quality minimum. The remaining 120 blocks supplied 5,903 retained events.

\section{Estimator and calibration implementation}

\subsection{Signal-processing and scoring implementation details}

Sentence reconstruction used EEGLAB marker pairs $10\rightarrow11$ for ordinary reading and $12\rightarrow13$ for reading followed by a control question. Response marker 15 was required to fall outside the sentence window. One-based EEGLAB latencies were converted once by $s_0=\operatorname{round}(s_{\mathrm{EEGLAB}})-1$. Candidate boundaries had to be finite, positive, nearly integral, unique, strictly ordered, and consistent with the material count. The accepted file prefixes (\texttt{gip}, \texttt{bip}, and \texttt{oip}) were retained as provenance rather than interpreted as quality labels.

The minimum-duration smoother operates on a linked list of contiguous label runs. It repeatedly selects the shortest sub-threshold run, with earlier onset breaking equal lengths, and reassigns the complete run to the adjacent state with the greatest summed squared spatial correlation. A common neighbor label is selected directly. Otherwise topographic evidence determines the choice, with lower state index resolving an exact tie. Boundary runs use their single neighbor, newly contiguous equal-label runs are merged, and iteration ends when every run meets the threshold. At 500~Hz, 20, 30, and 40~ms correspond to 10, 15, and 20 samples.

\paragraph{Deterministic calibration-bank allocation and hashes:} The bank is balanced over joint strata rather than through sequential marginal adjustments. For each participant, acquisition order and duration are first ranked with original row order resolving ties, then divided into as many as four equal-frequency bins. A stratum $h$ is the joint tuple (block, control-question status, order quartile, duration quartile), with $n_h$ eligible events among $n$ total events. Its ideal bank count is $q_h=96n_h/n$. The algorithm assigns $\lfloor q_h\rfloor$ events to every stratum and distributes the remaining places to the largest fractional remainders. Equal remainders are ordered by a stable hash of participant, repeat index, the literal tag \texttt{calibration-remainder}, and the stratum tuple. Within each stratum, events are hash-ordered by participant, repeat index, the tag \texttt{calibration-row}, and sentence identifier. The first allocated events enter the bank and all others remain analysis targets. The primary analysis used repeat index 0. This fully specifies the joint allocation objective and its tie rules.

All stable integers use the same mapping. Values are converted with their Python string representations, joined by the ASCII colon, encoded as UTF-8, and hashed with SHA-256. The first eight digest bytes are interpreted as an unsigned little-endian integer $H\in\{0,\ldots,2^{64}-1\}$. For a target event, the excluded bank offset is $H(\text{seed},\text{participant},\text{sentence},\texttt{event-reference-exclusion})\bmod96$. The randomized-rank variate for component $c$ is $U=H(\text{seed},\text{participant},\text{sentence},c,\texttt{event-rank-randomizer})/2^{64}$. These rules make allocation, exclusion, and tie randomization bit-for-bit reproducible from the saved identifiers and analysis seed.

For a METER-Jump target, the implementation loads the 96-event participant bank, applies its transition-confidence threshold, and excludes one hash-selected bank event so that real and pseudo-target references each contain 95 events. For reference counts $\mathbf m_i$ and target counts $\mathbf n_i$ from source state $i$, joint target and reference rows are drawn as
\begin{align}
Q_i\mid\mathbf m_i&\sim\operatorname{Dirichlet}(\mathbf m_i+\tfrac12\mathbf 1),\nonumber\\
P_i\mid\mathbf n_i,Q_i&\sim\operatorname{Dirichlet}(\mathbf n_i+\tau_AQ_i),\qquad\tau_A=1.
\label{eq:supp-dirichlet}
\end{align}
If a target source row is empty, the implementation draws $P_i\sim\operatorname{Dirichlet}(\tau_AQ_i)$ and retains that source with its fixed reference weight; it is not dropped or renormalized away. A reference row with no observed outgoing jumps remains proper through the Jeffreys $1/2$ mass in $Q_i$. Reference source weights are proportional to outgoing reference counts, with equal $1/3$ weights used only if the entire reference contains no outgoing transitions.
Each bank event is scored against its own $N-1$ reference. Candidates are ranked by the study-specific matching distance
\begin{equation}
d(x,c)=\left(\frac{\log d_c-\log d_x}{s_{\log d}}\right)^2+
\left(\frac{o_c-o_x}{\operatorname{range}(o)}\right)^2+
4\mathbf 1[b_c\ne b_x]+\mathbf 1[q_c\ne q_x],
\label{eq:supp-matching}
\end{equation}
where $d$, $o$, $b$, and $q$ denote duration, acquisition order, block, and control-question status. At most 50 pseudo-events are retained after duration-quartile matching. The stored output includes raw divergence, matched-null moments, $Z$, randomized rank, posterior summaries, transition evidence, candidate count, and excluded-reference identity. No application event updates the calibration bank. Reproducing the two equations here makes Supplementary Material self-contained when compiled separately from the main manuscript.

For posterior draws $R_{x}^{(1)},\ldots,R_{x}^{(D)}$ of a target event $x$ and the mean $\mu_{0x}$ of its matched pseudo-event scores, the reported null-centered interval is
\begin{equation}
I_x=\left[Q_{.025}\{R_x^{(d)}\}_{d=1}^{D}-\mu_{0x},\;
Q_{.975}\{R_x^{(d)}\}_{d=1}^{D}-\mu_{0x}\right].
\label{eq:supp-null-centered-interval}
\end{equation}
This diagnostic interval propagates conditional target- and reference-table uncertainty within the fixed bank and state representation. It does not include uncertainty in matched-null moments, bank selection, map estimation, sequence assignment, or group sampling, and it is not obtained by inverting the randomized event test. Its nominal coverage is therefore an operating characteristic evaluated in simulation rather than a general confidence guarantee.

\medskip
\refstepcounter{algorithm}
\noindent\textbf{Algorithm \thealgorithm. METER-Jump scoring for one target event}\label{alg:meter-jump}
\par\smallskip
\noindent\fbox{\begin{minipage}{0.94\textwidth}\small
\textbf{Input:} target $x$; participant's fixed 96-event bank $B$; fixed maps, confidence threshold, $\tau_A=1$, and analysis seed.\\
\textbf{1.} Smooth the continuous label stream, slice half-open event windows, collapse self-runs, and retain a jump only when both endpoints lie inside the event and pass the bank-derived confidence rule.\\
\textbf{2.} Exclude one deterministic hash-selected member of $B$; aggregate the other 95 sequences into reference counts and fixed source weights.\\
\textbf{3.} Draw reference rows $Q_i$ and target rows $P_i$ from Eq.~\ref{eq:supp-dirichlet}; evaluate the weighted conditional-destination JS divergence for every joint draw.\\
\textbf{4.} Treat each bank event in turn as a pseudo-target and score it against the other 95 bank events using the same construction.\\
\textbf{5.} Restrict candidates to the target-duration quartile (global fallback if fewer than ten); when more than 50 remain, rank them with Eq.~\ref{eq:supp-matching} and retain 50.\\
\textbf{6.} Compute $Z=(R_x-\mu_{0x})/\sigma_{0x}$ and the randomized upper-tail rank $p_x=\{n_>+U(n_=+1)\}/(m+1)$, where $U$ is the stored SHA-256-derived variate.\\
\textbf{Output:} raw/posterior score summaries, $Z$, randomized rank, interval in Eq.~\ref{eq:supp-null-centered-interval}, evidence diagnostics, matches, and excluded-reference identity.
\end{minipage}}

Algorithm~\ref{alg:meter-jump} separates the fixed construction choices from the event-specific calculations and makes explicit that a target never updates its participant bank.

Duration is the opportunity variable because it is defined before sequence construction and captures time available to observe transitions without conditioning on realized jump count, which can itself respond to the transition process. In the 96-event bank, a duration quartile usually held fewer than 50 eligible candidates, so all were retained and the categorical penalties did not change the primary target coordinates. The numerical terms become active only when the eligible set exceeds the cap; Table~\ref{tab:matching-audit} quantifies this behavior. The current application therefore implements participant-specific duration-stratified calibration rather than direct matching on jump count, source-row evidence, or active multivariable context matching.

The auxiliary order-selection analysis compared first- and second-order jump prediction by balanced two-fold cross-validation. If $G$ denotes the second-minus-first-order log-score gain in nats per jump, the prespecified simulation anchor was
\begin{equation}
\delta_{\mathrm{order}}=\max\!\left\{Q_{0.95}(G\mid\lambda_2=0),\;0.10\,\operatorname{median}(G\mid\lambda_2=0.8)\right\}=0.0349.
\label{eq:supp-order-threshold}
\end{equation}
The observed gain was 0.00205 nats per jump, below both the null-tail and moderate/strong simulated-effect scales. Equation~\ref{eq:supp-order-threshold} is an auxiliary complexity-selection rule, not a universal physiological threshold.

Every run stored the resolved configuration, software environment, source and input hashes, progress records, participant- or fold-level outputs, and summary tables. Resume validation checked data identity, configuration, replicate sequence, and expected cell counts before continuing an interrupted run. The final regression suite contained 88 passing tests. On the recorded Linux x86\_64 workstation (32 logical CPUs; single-threaded BLAS), the primary preprocessing and map-estimation run required 1,952~s (32.5~min). The primary METER scoring run required 3,909~s (65.2~min) with 18 workers. These wall-clock measurements exclude dataset transfer and figure generation and should be read as machine-specific reproducibility information, not benchmark guarantees.

The released Cz signal was exactly zero before added processing. CAR included it with the other 104 signals; in the no-additional-CAR sensitivity it remained zero. Both streams retained the same 105-dimensional geometry, and every polarity-invariant map correlation mean-centered each topography before normalization.

\subsection{Continuous-map uncertainty sampling and alignment}

The continuous-sample analysis used two acquisition-spaced valid blocks per participant, selected at equally spaced positions in the ordered block list. A 20-s portion of each released cleaned recording was centered within the block and received the primary zero-phase 1--40-Hz filter and CAR. Discarding 5~s from each filtered edge left a central 4-s segment with a 30-sample smoothing pad. The temporal comparison therefore included 36 segments from all 18 participants.

Within each selected block, GFP was normalized and divided into low, middle, and high tertiles. Twenty-five topographies were sampled without replacement from each tertile, and 25 GFP peaks separated by at least two samples formed a fourth stratum. Across participants and blocks, each stratum contributed 900 topographies. For each of 500 participant bootstraps, participant map sets were sampled with replacement, aligned to the fixed $K=3$ state identities by Hungarian matching on absolute spatial correlation, polarity-aligned, averaged with equal participant weight, and normalized state by state. The resulting consensus maps labeled all 3,600 stratified topographies before smoothing and the padded continuous segments before and after the same 30-ms run-merging rule used in the primary analysis. Disagreement was the fraction of labels differing from the fixed-map assignment; temporal disagreement was Levenshtein distance between run-collapsed chains divided by the longer chain.

This consensus construction is computationally lighter than refitting participant and group maps for every bootstrap. An explicit approximation check used 100 identical participant-bootstrap samples and compared each consensus result with a complete 25-start, equal-participant-weight group refit. The median minimum matched-state correlation was .968 and its 10th percentile was .898. The check therefore supports the approximation for this diagnostic, while its scope remains narrower than a full sensor-level error model or within-participant peak-resampling analysis.

\subsection{Application implementation details}

Surprisal was computed in bits with \url{openai-community/gpt2}, snapshot \texttt{607a30d7}, Transformers 4.55.4, and PyTorch 2.10. Subword values were combined into whitespace-delimited words. With no preceding-sentence context, the first word was coded as zero; a sensitivity divided summed surprisal by $n-1$ and excluded that word. Dependency measures used spaCy's small English model (\texttt{en\_core\_web\_sm}, version 3.8.0), with its transformer model (\texttt{en\_core\_web\_trf}, version 3.8.0) as the parser sensitivity; readability used textstat. These sensitivities preserved the primary regression specification and multiplicity rule.

Numerically, excluding the first word from mean surprisal changed no controlled METER-Jump coefficient by more than .00335; its minimum max-$T$ $p$ was .2151, and no corrected decision changed. Replacing the small spaCy parser with the transformer parser changed no coefficient by more than .00408; its minimum max-$T$ $p$ was .2678, again with no decision change. These checks support decision stability for the two evaluated feature-extraction choices, not equivalence of every coefficient or parser.

\begin{figure*}[htpb]
\centering
\includegraphics[width=0.82\textwidth]{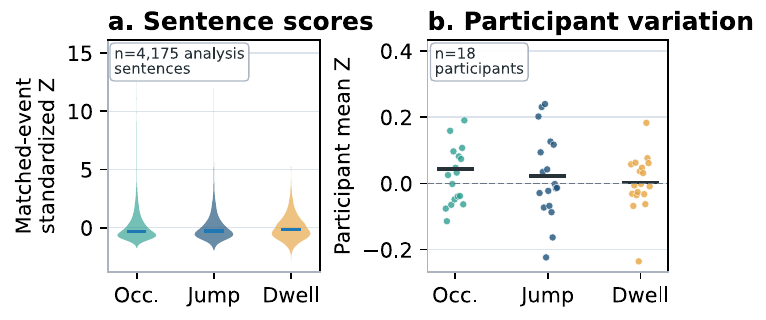}
\caption{Descriptive event-score diagnostics. (a) Event-level coordinates for occupancy, jump destination, and exploratory dwell reconfiguration. (b) Participant means show between-reader variation. These distributions are descriptive; group inference uses equal-participant slopes. Finite-event uncertainty and the component-calibration comparison are paired in main-text Fig.~5}
\label{fig:scores}
\end{figure*}

\begin{figure*}[htpb]
\centering
\includegraphics[width=0.8\textwidth]{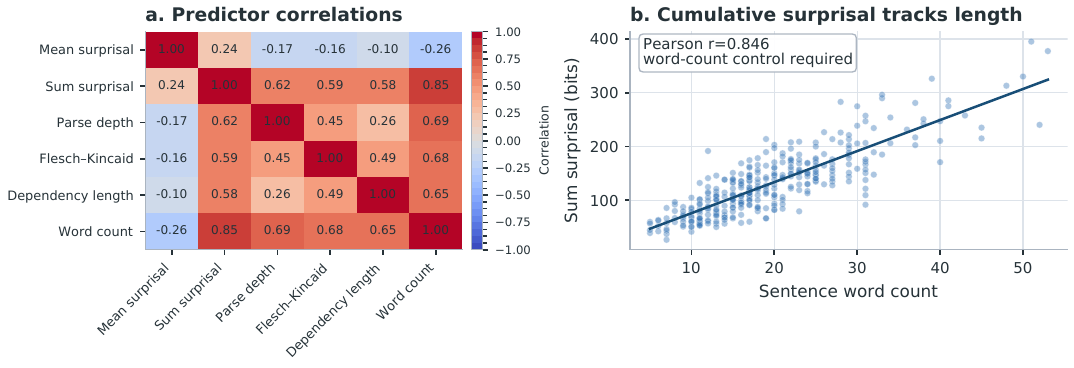}
\caption{Linguistic predictor diagnostics. (a) Correlations among the five predefined sentence measures and word count. (b) Summed GPT-2 surprisal is strongly related to word count ($r=0.846$), motivating explicit word-count adjustment in the participant models}
\label{fig:predictors}
\end{figure*}

The supplementary NR/TSR control retained 11,256 of 11,824 cached rows from 16 participants with structurally complete inputs for both tasks. Within each crossed fold, the 30\% participant-specific calibration subset selected without task labels formed the reference bank and was excluded from classification; after that exclusion, 7,843 unique held-out rows contributed to the participant-specific evaluation. Its 80 outer folds crossed each held participant with five content-hash text folds. Maps, vocabulary, imputation, and scaling were trained within each fold. Text features were lowercase word unigrams and bigrams with sublinear term frequency, smoothed inverse document frequency, minimum document frequency two, at most 20,000 features, and $\ell_2$ normalization. A linear support-vector classifier with $C=1$ compared text alone, classical microstate features, and nested additions of occupancy and METER-Jump.

\begin{figure*}[htpb]
\centering
\includegraphics[width=0.8\textwidth]{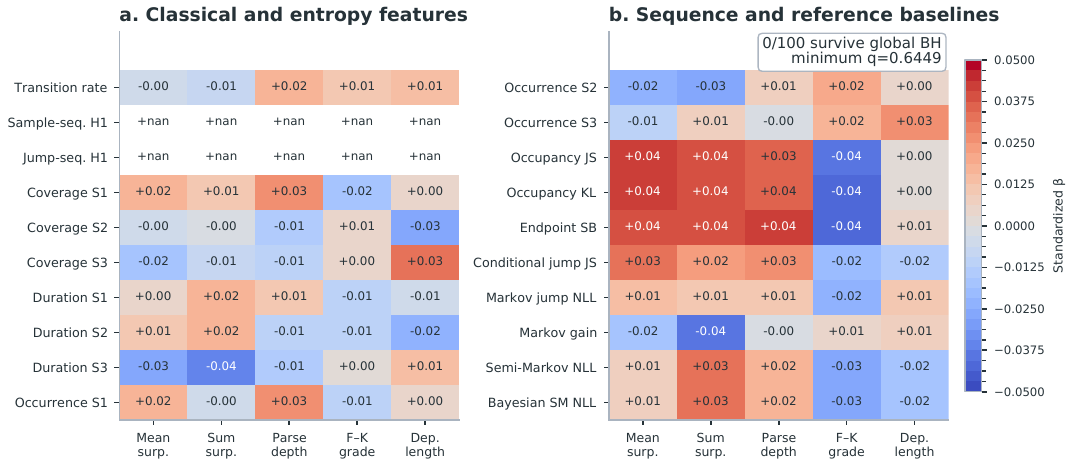}
\caption{Complete controlled baseline--predictor coefficient matrix. Classical summaries appear in (a) and reference/sequence measures in (b). Labels are publication names rather than implementation column names. No association remains after BH correction over the expanded 100-test table (minimum $q=0.7166$). Entropy-rate rows are reported numerically in Table~\ref{tab:entropy-results}}
\label{fig:supp-baseline-associations}
\end{figure*}

Figure~\ref{fig:scores} summarizes the event-level coordinates and their variation across participants; main-text Fig.~5 shows how METER-Jump posterior uncertainty declines as retained transition evidence grows. Figure~\ref{fig:predictors} documents the predictor correlation structure, including the strong association between summed surprisal and word count that motivated explicit word-count adjustment. Figure~\ref{fig:supp-baseline-associations} then reports the complete controlled baseline screen: no classical, entropy-rate, sequence, or reference-relative comparator survived the predefined 100-test BH correction.

%
\suppressfloats[t]

Table~\ref{tab:associations} reports the complete controlled naturalistic-reading estimates corresponding to main-text Fig.~10. Several uncorrected tendencies and participant heterogeneity are visible, but no jump, occupancy, or exploratory dwell predictor survives its five-predictor max-$T$ family.

\begin{center}
\captionof{table}{Controlled Naturalistic-Reading Associations}
\label{tab:associations}
\centering
\scriptsize
\setlength{\tabcolsep}{3.5pt}
\begin{tabular}{llrrrrr}
\toprule
\textbf{Outcome role} & \textbf{Predictor} & $\boldsymbol{\beta}$ & \shortstack{\textbf{$t$-summary}\\\textbf{95\% interval}} & \textbf{Exact $p$} & \textbf{Max-$T$ $p$} & \shortstack{\textbf{Positive}\\\textbf{participants}} \\
\midrule
METER-Jump (primary) & Mean surprisal & 0.0154 & [$-0.0116$, 0.0425] & 0.2434 & 0.6954 & 13/18 \\
 & Sum surprisal & 0.0113 & [$-0.0384$, 0.0609] & 0.6367 & 0.9869 & 11/18 \\
 & Parse-tree depth & 0.0252 & [$-0.0003$, 0.0507] & 0.0540 & 0.2124 & 12/18 \\
 & Flesch--Kincaid & $-0.0181$ & [$-0.0568$, 0.0206] & 0.3446 & 0.8238 & 9/18 \\
 & Dependency length & $-0.0210$ & [$-0.0543$, 0.0124] & 0.2031 & 0.6186 & 7/18 \\
\midrule
METER-Occ (comparator) & Mean surprisal & 0.0392 & [$-0.00003$, 0.07841] & 0.0493 & 0.1796 & 13/18 \\
 & Sum surprisal & 0.0375 & [$-0.0279$, 0.1028] & 0.2579 & 0.7199 & 11/18 \\
 & Parse-tree depth & 0.0281 & [$-0.0178$, 0.0740] & 0.2131 & 0.6637 & 11/18 \\
 & Flesch--Kincaid & $-0.0297$ & [$-0.0753$, 0.0159] & 0.1896 & 0.6056 & 8/18 \\
 & Dependency length & $-0.0027$ & [$-0.0790$, 0.0737] & 0.9526 & 1.0000 & 8/18 \\
\midrule
Dwell (exploratory) & Mean surprisal & $-0.0003$ & [$-0.0145$, 0.0139] & 0.9623 & 1.0000 & 9/18 \\
 & Sum surprisal & $-0.0071$ & [$-0.0381$, 0.0239] & 0.6308 & 0.9896 & 8/18 \\
 & Parse-tree depth & 0.0285 & [0.0018, 0.0552] & 0.0388 & 0.1633 & 12/18 \\
 & Flesch--Kincaid & $-0.0183$ & [$-0.0508$, 0.0142] & 0.2523 & 0.7139 & 6/18 \\
 & Dependency length & $-0.0241$ & [$-0.0502$, 0.0020] & 0.0689 & 0.2728 & 6/18 \\
\bottomrule
\end{tabular}
\begin{flushleft}\footnotesize Intervals use the participant-slope $t$ summary and are descriptive compatibility intervals; exact $p$-values enumerate sign flips. The intervals are not obtained by inversion of the exact test, so zero crossing and exact-test significance need not coincide.\end{flushleft}
\end{center}

\begin{figure*}[htpb]
\centering
\includegraphics[width=0.8\textwidth]{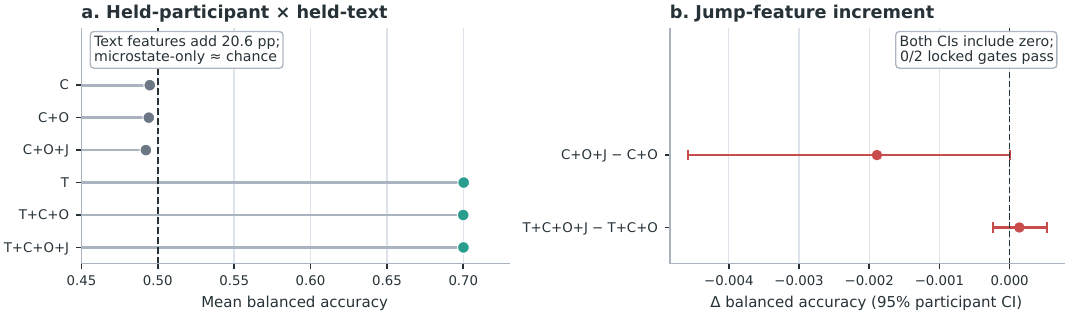}
\caption{Supplementary participant-specific held-participant/held-text NR/TSR control. (a) Text features drive balanced accuracy. (b) Paired METER-Jump increments and participant-bootstrap intervals include zero. The analysis evaluates incremental prediction when the held participant supplies an unlabeled participant-specific calibration bank}
\label{fig:prediction}
\end{figure*}

Figure~\ref{fig:prediction} and Table~\ref{tab:prediction} provide the separate predictive control: text features discriminate NR from TSR, whereas adding METER-Jump produces no reliable increment beyond either conventional microstate features or the text model.

\begin{table*}[htpb]
\caption{Personalized Held-Participant/Held-Text NR/TSR Prediction}
\label{tab:prediction}
\centering
\begin{tabular}{lrr}
\toprule
\textbf{Model} & \textbf{Balanced accuracy [95\% CI]} & \textbf{AUROC [95\% CI]} \\
\midrule
$C$ & .4948 [.4843, .5000] & .4682 [.4311, .5006] \\
$C+O$ & .4941 [.4823, .5003] & .4608 [.4298, .4890] \\
$C+O+J$ & .4922 [.4790, .5001] & .4587 [.4298, .4853] \\
$T$ & .7002 [.6952, .7051] & .7635 [.7574, .7700] \\
$T+C+O$ & .6999 [.6950, .7050] & .7635 [.7573, .7701] \\
$T+C+O+J$ & .7001 [.6950, .7051] & .7634 [.7572, .7700] \\
\bottomrule
\end{tabular}
\begin{flushleft}
\footnotesize Participant means and 20,000-draw participant-bootstrap intervals. $C$: classical microstate features; $O$: occupancy reconfiguration; $J$: METER-Jump; $T$: word TF--IDF. The held participant supplies only an unlabeled participant-specific calibration bank, not classifier training data.
\end{flushleft}
\end{table*}

\begin{center}
\captionof{table}{Leave-one-participant-out analysis of primary METER-Jump group means}
\label{tab:loo-slopes}
\centering
\small
\begin{tabular}{lrr}
\toprule
\textbf{Predictor} & \textbf{Full mean} & \textbf{LOO range} \\
\midrule
Mean surprisal & .0154 & [.0107, .0229] \\
Sum surprisal & .0113 & [.0030, .0239] \\
Parse-tree depth & .0252 & [.0203, .0305] \\
Flesch--Kincaid & $-.0181$ & [$-.0243$, $-.0072$] \\
Dependency length & $-.0210$ & [$-.0289$, $-.0127$] \\
\bottomrule
\end{tabular}
\begin{flushleft}\footnotesize Each range contains 18 equal-weight means, each omitting one participant from the saved controlled participant slopes. All retained the full-sample direction; this is a descriptive influence analysis, not a new hypothesis test.\end{flushleft}
\end{center}

Table~\ref{tab:loo-slopes} shows that omitting any single participant preserves the direction of all five primary group means, although this descriptive stability does not replace corrected inference.

\begin{landscape}
\subsection{Predictive and conventional controls}

Table~\ref{tab:baseline-specification} records the estimand, reference construction, regularization, censoring, and orientation of each sequence-level comparator. This specification is necessary because the methods are distinct measurement formulations rather than scores produced under a common reference design.

\begin{center}
\captionof{table}{Implementation Specification of Sequence-Level Comparators}
\label{tab:baseline-specification}
\centering
\footnotesize
\setlength{\tabcolsep}{4pt}
\begin{tabularx}{\linewidth}{p{.14\linewidth}p{.17\linewidth}p{.15\linewidth}p{.22\linewidth}X}
\toprule
\textbf{Comparator} & \textbf{Estimand / score} & \textbf{Reference} & \textbf{Regularization and censoring} & \textbf{Orientation and scope} \\
\midrule
Occupancy JS & Marginal state-prevalence JS & Target-excluded participant LOO & Jeffreys $1/2$ mass; no dwell & Larger means greater prevalence change. \\
Sample-sequence / jump-sequence first-order entropy rate & First-order conditional entropy & None; event sequence & Plug-in event counts; not Sample Entropy (SampEn) & Larger means greater sequence uncertainty; complementary comparator. \\
Endpoint SB & Minimum endpoint bridge cost & Target-excluded participant LOO & Converged KL projection to target/reference endpoints & Larger means greater endpoint redistribution; does not establish changed conditional jumps. \\
Conditional jump JS & Source-weighted destination JS & Target-excluded participant LOO & Jeffreys $1/2$ off-diagonal mass & Larger means greater plug-in conditional jump change. \\
Markov NLL / gain & Held-out jump destination loss / gain over occupancy & Balanced two-fold within participant & Jeffreys $1/2$ transition mass & Lower NLL is better; positive gain favors jump history. \\
Discrete semi-Markov & Joint jump and dwell NLL per event & Balanced two-fold within participant & Common 11-bin dwell grid; right-censor survival likelihood & Lower held-out NLL is better. \\
Hierarchical Bayesian semi-Markov & Posterior-predictive joint NLL per event & Balanced two-fold; other-participant population prior & Transition and dwell concentration 4; same censoring and bins & Lower held-out NLL is better; common-representation comparator, not external-package reproduction. \\
\bottomrule
\end{tabularx}
\begin{flushleft}\footnotesize All comparators use the saved smoothed $K=3$ labels and the same analysis observations; maps are not re-estimated from linguistic or task outcomes.\end{flushleft}
\end{center}
\end{landscape}

The archived Barzon transport-cost analysis was reproduced from the public v0.0.1 repository. Legacy costs agreed to a maximum absolute error of $3.8\times10^{-15}$, the converged solver correlated 0.99996 with the archived values, and the recomputed inferential statistics matched the notebook's printed four-decimal values to within $4.1\times10^{-5}$. The reproduction comprised 264 condition rows from 44 participants. This check validates the comparator implementation and archived inference; it is not an external replication of the present ZuCo result.

Table~\ref{tab:conventional-controls} reports a separate public ZuCo NR/TSR benchmark using conventional EEG, eye-movement, and text features under the same 16-participant by five-content-fold outer split. It used the full matched public benchmark before the participant-specific calibration exclusion and therefore must not be compared row-for-row with Table~\ref{tab:prediction}. The result is useful as a modality check: gaze and text contain clear task information. Simple EEG summaries have balanced accuracy near chance, although several AUROCs are modestly above .5, so they should not be described as wholly uninformative.

\begin{table}[htpb]
\caption{Separate Conventional NR/TSR Controls}
\label{tab:conventional-controls}
\centering
\begin{tabular}{lcc}
\toprule
\textbf{Feature set} & \textbf{Balanced accuracy} & \textbf{AUROC} \\
\midrule
Alpha mean & .5072 & .6251 \\
Beta mean & .5041 & .5771 \\
Gamma mean & .4960 & .5172 \\
Theta mean & .5198 & .5914 \\
EEG channel means & .5139 & .5825 \\
All electrode features & .5384 & .5893 \\
Sentence gaze/saccade & .6140 & .6974 \\
Word TF--IDF & .6998 & .7711 \\
\bottomrule
\end{tabular}
\begin{flushleft}\footnotesize Eighty crossed participant--content folds. The all-electrode feature set had 8,755 test rows; the other physiological sets had approximately 11,323 rows. These are separate conventional controls, not feature ablations on the 7,843 participant-specific validation rows.\end{flushleft}
\end{table}

\begin{figure*}[htpb]
\centering
\includegraphics[width=0.8\textwidth]{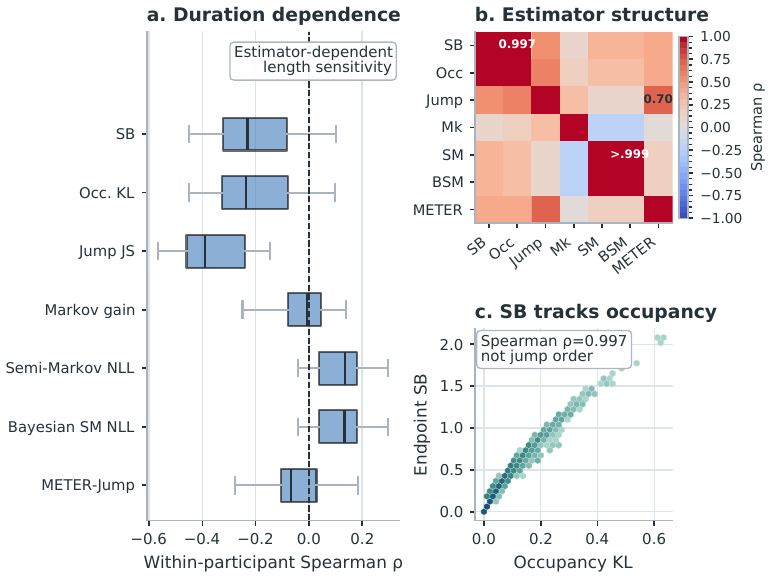}
\caption{Complementary-method diagnostics. (a) Duration dependence. (b) Rank relationships among estimators; annotations are restricted to the relationships discussed in the text or $|\rho|\geq0.65$. (c) The evaluated endpoint Schr\"odinger bridge and occupancy KL are nearly order-equivalent. The panels document that the methods estimate non-interchangeable properties rather than forming a winner--loser ranking}
\label{fig:supp-comparator-diagnostics}
\end{figure*}

Figure~\ref{fig:supp-comparator-diagnostics} places the complementary estimators on a common descriptive scale. Duration dependence and the strong endpoint-bridge--occupancy association reinforce that these measures cannot be interpreted as interchangeable tests of jump reconfiguration.

\section{Operating-characteristic and robustness evidence}
\suppressfloats[t]

Tables~\ref{tab:type1}--\ref{tab:simulation-dgp} define the central simulation evidence. Table~\ref{tab:type1} confirms near-nominal event-null error across both seeds and all four regimes; Table~\ref{tab:component-ablation} shows that posterior integration or shrinkage without matched-event calibration does not yield a transferable global cutoff; and Table~\ref{tab:simulation-dgp} makes the generator interventions and their inferential purposes explicit. Together they support calibration for the evaluated mechanisms, not a distribution-free guarantee.

\begin{center}
\captionof{table}{Event-Null False-Positive Rates}
\label{tab:type1}
\centering
\small
\begin{tabular}{lccc}
\toprule
\textbf{Null regime} & \textbf{Seed 20260829} & \textbf{Seed 20260830} & \shortstack{\textbf{Misspecification}\\\textbf{study}} \\
\midrule
Participant fitted & .046 [.035, .061] & .047 [.036, .062] & -- \\
Unequal occupancy & .050 [.038, .065] & .056 [.043, .072] & -- \\
Rare transition & .044 [.033, .059] & .052 [.040, .068] & -- \\
GFP-peak bootstrap maps & .049 [.037, .064] & .054 [.042, .070] & -- \\
Time-varying transitions & -- & -- & .054 [.037, .077] \\
Destination-dependent dwell & -- & -- & .060 [.042, .084] \\
\bottomrule
\end{tabular}
\begin{flushleft}\footnotesize Rates with 95\% Wilson intervals. The two principal seeds used 1,000 replicates per regime. Each separate misspecification condition used 500 replicates.\end{flushleft}
\end{center}

\begin{table}[htpb]
\caption{Paired Estimator-Component Ablation}
\label{tab:component-ablation}
\centering
\small
\textbf{(a) False-positive rate by null regime}\\[3pt]
\begin{tabular}{lrrrr}
\toprule
\textbf{Variant} & \textbf{Fitted} & \textbf{Unequal} & \textbf{Rare} & \textbf{Peak-map} \\
\midrule
Plug-in conditional JS & .058 & .176 & .004 & .206 \\
Shrinkage-only JS & .048 & .130 & .006 & .158 \\
Posterior-integrated JS & .060 & .190 & .004 & .234 \\
METER-Jump & .055 & .045 & .055 & .045 \\
\bottomrule
\end{tabular}
\\[8pt]
\textbf{(b) Detection probability under the defined alternative}\\[3pt]
\begin{tabular}{lrr}
\toprule
\textbf{Variant} & \textbf{Fixed maps} & \textbf{Peak-map perturbation} \\
\midrule
Plug-in conditional JS & .562 & .572 \\
Shrinkage-only JS & .706 & .600 \\
Posterior-integrated JS & .572 & .616 \\
METER-Jump & .680 & .226 \\
\bottomrule
\end{tabular}
\begin{flushleft}\footnotesize Uncalibrated cutoffs were fitted on 500 developmental participant-null replicates and evaluated on 500 paired replicates. METER null rates use all 1,000 native matched-event tests; power uses the common 500-replicate evaluation half. Across the four null regimes, absolute Spearman correlations between event duration and the globally thresholded component scores ranged from .44 to .78; the duration correlations for METER-Jump ranged from $-.091$ to .024.\end{flushleft}
\end{table}

\begin{figure}[htpb]
\centering
\includegraphics[width=0.52\textwidth]{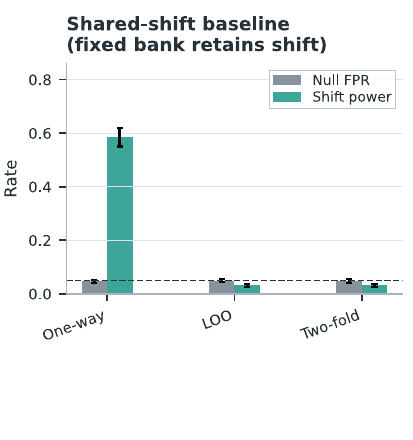}
\caption{Reference-design comparison for a shared shift relative to a predesignated bank. Target-disjoint, leave-one-out (LOO), and balanced two-fold formulations have similar null rates, but the LOO and two-fold reference sets contain the same simulated shift and therefore absorb it. This result is specific to the simulated fixed-baseline estimand}
\label{fig:supp-reference-design}
\end{figure}

Figure~\ref{fig:supp-reference-design} isolates the consequence of allowing shifted analysis events to enter their own comparison reference. The attenuated LOO and two-fold alternatives include the same simulated shift in their reference sets; they are not failed versions of the fixed-bank estimand.

\begin{table*}[htpb]
\caption{Participant-Fitted Simulation Data-Generating Processes}
\label{tab:simulation-dgp}
\centering
\footnotesize
\begin{tabularx}{\textwidth}{p{0.18\textwidth}X X}
\toprule
\textbf{Regime} & \textbf{Generator intervention} & \textbf{Purpose} \\
\midrule
Participant-fitted null & $A_u$ and fitted state dwell means are unchanged for bank and target. & Baseline type-I behavior at observed event lengths. \\
Unequal-occupancy null & The mean dwell of state 1 is multiplied by 4 while $A_u$ is unchanged for both bank and target. & Strong sample-occupancy imbalance without a target effect. \\
Rare-transition null & Every incoming edge to state 3 is multiplied by 0.08, the diagonal remains zero, and rows are renormalized for both bank and target. & Sparse destination counts without a target effect. \\
GFP-peak-derived map null & Baseline labels are reassigned from a sampled GFP-peak-derived bootstrap-map correlation bank and then smoothed for both bank and target. & Assignment error under the implemented peak-derived null. \\
Time-varying-transition null & At jump fraction $f$, the transition row is $(1-f)A_u+fA_u^*$ followed by row normalization, where $A_u^*$ is an occupancy-coupled transformation of $A_u$ with strength .8. The same within-event drift generates bank and target sequences. & Departure from time-homogeneous first-order dynamics without a bank--target effect. \\
Destination-dependent-dwell null & Each source-state dwell mean is multiplied by $.65$, $1.00$, or $1.35$ according to the destination. Each source row is then rescaled so its transition-weighted dwell mean equals the participant's original fitted mean; durations are bounded below by one sample. The same rule generates bank and target sequences. & Dependence between dwell and destination while preserving source-level mean duration. \\
Jump alternative & For stationary jump visitation $\nu_u$ and flux $F_{ij}=\nu_{ui}A_{uij}$, let $\epsilon=\lambda\min(F_{02},F_{10},F_{21})$, $\lambda=0.8$. Add $\epsilon$ to $0\!\to\!1$, $1\!\to\!2$, and $2\!\to\!0$ and subtract it from the reverse-cycle edges; set $A^{(\lambda)}_{ij}=F^{(\lambda)}_{ij}/\nu_{ui}$. & Changes destinations while preserving $\nu_uA^{(\lambda)}=\nu_u$. \\
\bottomrule
\end{tabularx}
\begin{flushleft}\footnotesize At $\lambda=.8$, the participant-specific perturbations had a median largest absolute off-diagonal probability change of .280 (IQR .243--.315) and a median stationary-weighted JS change of .0349 nats (IQR .0279--.0469). These summaries translate $\lambda$ into changes in the fitted transition matrices; they do not make it a physiological effect-size unit.\end{flushleft}
\end{table*}

\begin{table}[htpb]
\caption{Continuous Map-Uncertainty Analysis}
\label{tab:continuous-map}
\centering
\begin{tabular}{lcc}
\toprule
\textbf{GFP stratum} & \textbf{Before smoothing} & \textbf{After smoothing} \\
\midrule
Low & .1086 & .1772 \\
Mid & .0997 & .1545 \\
High & .0872 & .1174 \\
Peak & .0962 & --- \\
\bottomrule
\end{tabular}
\begin{flushleft}\footnotesize Mean disagreement over 500 bootstrap consensus maps. Peak is defined only before temporal smoothing; mean post-smoothing jump-chain edit distance was 0.1913.\end{flushleft}
\end{table}

Table~\ref{tab:continuous-map} complements the isolated GFP-peak experiment by analyzing continuous samples. Disagreement is greatest in the low-GFP stratum and increases after smoothing, supporting the claim that reproducible group maps do not imply error-free dynamic label sequences.

\begin{table*}[htpb]
\caption{Common-Target Fixed-Map Reference-Size Operating Characteristics ($\lambda=0.8$ Stationary-Flux Perturbation)}
\label{tab:calibration}
\centering
\scriptsize
\begin{tabular}{rccccccc}
\toprule
\shortstack{\textbf{Reference}\\\textbf{events}} & \shortstack{\textbf{Matched}\\\textbf{pseudo-events}} & \shortstack{\textbf{Null FPR}\\\textbf{range}} & \shortstack{$\boldsymbol{\lambda=.8}$\\\textbf{detection}} & \shortstack{\textbf{Median}\\$\boldsymbol{|\mathrm{bias}|}$} & \shortstack{\textbf{Median interval}\\\textbf{width}} & \shortstack{\textbf{Event-score}\\\textbf{coverage}} & \shortstack{\textbf{Sparse-row}\\\textbf{events}} \\
\midrule
32$^{a}$ & 32 & .040--.059 & .526 & .0121 & .0579 & .886 & .014 \\
48  & 12 & .049--.053 & .472 & .0120 & .0585 & .926 & .014 \\
64  & 16 & .049--.056 & .584 & .0120 & .0579 & .923 & .014 \\
80  & 20 & .046--.059 & .663 & .0114 & .0588 & .921 & .014 \\
\textbf{96}  & 24 & .040--.064 & .673 & .0114 & .0578 & .915 & .014 \\
120 & 30 & .047--.055 & .673 & .0116 & .0579 & .907 & .014 \\
144 & 36 & .049--.056 & .681 & .0115 & .0578 & .916 & .014 \\
\bottomrule
\end{tabular}
\begin{flushleft}\footnotesize $^{a}$The 32-event bank invoked the predefined global-matching fallback because fewer than ten candidates were available within some duration quartiles. All seven rows use the same 1,000 common simulated targets. The primary two-seed 96-event estimates were detection .687 and coverage .918; the later common-target analysis did not alter the application. Sparse rows describe target information and therefore do not vary with reference size.\end{flushleft}
\end{table*}

Table~\ref{tab:calibration} shows the practical reference-size plateau under the fixed-map $\lambda=0.8$ alternative. The 80--96-event range approaches the largest evaluated detection probability, whereas the 32-event row invokes a different global-matching regime and should not be read as a pure sample-size comparison.

\begin{table*}[htpb]
\caption{Event-Matching Sensitivity Analysis}
\label{tab:matching-audit}
\centering
\scriptsize
\begin{tabular}{lccccc}
\toprule
\textbf{Matcher} & \shortstack{\textbf{Pseudo-event rejection}\\\textbf{rate [95\% CI]}} & $\boldsymbol{\rho(Z,Z_0)}$ & $\boldsymbol{\rho(Z,d)}$ & \textbf{Block match} & \textbf{Question match} \\
\midrule
Quartile, $N=8$ & .0434 [.0348,.0541] & .912 & $-.039$ & .378 & .846 \\
Quartile, $N=12$ & .0440 [.0353,.0547] & .957 & $-.040$ & .265 & .838 \\
Quartile, $N=16$ & .0492 [.0400,.0604] & .978 & $-.055$ & .208 & .814 \\
Quartile, $N=25$ & .0451 [.0363,.0560] & 1.000 & $-.051$ & .149 & .782 \\
Quartile, $N=50$ & .0451 [.0363,.0560] & 1.000 & $-.051$ & .149 & .782 \\
Weak penalty $(2,.5)$ & .0492 [.0400,.0604] & 1.000 & $-.051$ & .149 & .782 \\
Strong penalty $(8,2)$ & .0475 [.0384,.0585] & 1.000 & $-.051$ & .149 & .782 \\
Global, $N=50$ & .0411 [.0327,.0515] & .896 & $-.312$ & .259 & .847 \\
\bottomrule
\end{tabular}
\begin{flushleft}\footnotesize Matching sensitivity analysis using the production squared-distance metric. Each of 1,728 calibration events is excluded from its own candidate reference. $Z_0$: specified full within-quartile matcher; $d$: duration. No quartile matcher used the global fallback. With $N=50$, each duration quartile had fewer than 50 eligible candidates, so all were retained and changing the categorical penalties did not change target coordinates.\end{flushleft}
\end{table*}

Table~\ref{tab:matching-audit} shows why duration-conditioned matching was retained: the internal pseudo-event rejection rate remains near 0.05 and target scores remain closely aligned with the primary matcher, whereas global matching is more duration-dependent. It also shows why the numerical penalty choice is immaterial in this 96-event design: the 50-neighbor cap retains every candidate in the eligible duration quartile. The $N=16$ row demonstrates that a smaller, context-active neighborhood preserves pseudo-event exchangeability and closely tracks the primary coordinate, but the five application regressions were not rerun under that alternative. Larger reference banks should re-evaluate the context terms directly.

\begin{table}[htpb]
\caption{Score-Draw Convergence}
\label{tab:draw-convergence}
\centering
\scriptsize
\begin{tabular}{lcc}
\toprule
\textbf{Retained draws} & \textbf{Group-$\beta$ range} & \textbf{Minimum max-$T$ $p$} \\
\midrule
16 & [$-.0162$, .0279] & .152 \\
32 & [$-.0228$, .0262] & .158 \\
64 & [$-.0253$, .0256] & .168 \\
128 & [$-.0253$, .0277] & .130 \\
\bottomrule
\end{tabular}
\begin{flushleft}\footnotesize Five primary jump-predictor slopes. The maximum absolute group-slope change between 32 and 128 draws was .00254; no retained-draw setting changed the corrected conclusion.\end{flushleft}
\end{table}

Table~\ref{tab:draw-convergence} verifies that 32 retained score draws are sufficient for the reported group analysis: increasing the count to 128 changes no corrected conclusion and only minimally changes the five primary slopes.

\begin{table*}[htpb]
\caption{End-to-End Processing Sensitivity of METER-Jump}
\label{tab:end-to-end-processing}
\centering
\scriptsize
\setlength{\tabcolsep}{3.5pt}
\begin{tabular}{lrrrrrrrr}
\toprule
\textbf{Stream} & \shortstack{\textbf{Common}\\\textbf{events}} & \shortstack{$\boldsymbol{\rho}$\textbf{,}\\$\boldsymbol{Z}$} & \shortstack{\textbf{Median}\\$\boldsymbol{\rho_u}$} & \shortstack{\textbf{Median}\\$\boldsymbol{|\Delta Z|}$} & \shortstack{\textbf{Pseudo-event}\\\textbf{rejection rate}} & \shortstack{\textbf{Median/max}\\$\boldsymbol{|\Delta\beta|}$} & \shortstack{\textbf{Minimum}\\\textbf{max-}$\boldsymbol{T}$ $\boldsymbol{p}$} & \shortstack{\textbf{Decision}\\\textbf{changed}} \\
\midrule
2--20 Hz + CAR & 4,167 & .330 & .309 & .634 & .0532 & .0065/.0229 & .4557 & No \\
20 ms smoothing & 4,175 & .556 & .567 & .496 & .0492 & .0075/.0232 & .0082 & Yes \\
40 ms smoothing & 4,175 & .633 & .618 & .432 & .0498 & .0093/.0140 & .0539 & No \\
No additional CAR & 4,167 & .986 & .988 & .039 & .0469 & .0034/.0042 & .2937 & No \\
\bottomrule
\end{tabular}
\begin{flushleft}\footnotesize Every alternative rebuilt continuous labels where applicable, the participant bank selected without application outcomes, and all scores while holding the estimator, bank size, posterior resolution, matching rule, and correction hierarchy fixed. $\rho$: global Spearman correlation of METER-Jump $Z$ with the primary 30-ms analysis; $\rho_u$: within-participant Spearman correlation. $|\Delta\beta|$ summarizes changes in the five controlled jump coefficients. The pseudo-event rejection rate comes from an internal leave-one-out calibration-bank exchangeability analysis, not a processing-specific participant-fitted simulation or power estimate. Sensitivity $p$-values retain within-analysis max-$T$(5) correction but are not corrected across processing variants. ``Decision changed'' denotes whether any corrected jump conclusion differed from the primary analysis.\end{flushleft}
\end{table*}

Table~\ref{tab:end-to-end-processing} separates numerical agreement, internal pseudo-event rejection, coefficient movement, and decision stability across four processing sensitivity analyses. The corresponding coefficient trajectories and passband decomposition are displayed in the main manuscript's processing-sensitivity figure. A near-0.05 internal rejection rate does not imply coordinate invariance: the 20-ms definition changes one within-analysis decision, while the 2--20-Hz analysis substantially changes $Z$ without changing the corrected conclusion.

\begin{table*}[htpb]
\caption{First-Order Entropy-Rate Application Comparators}
\label{tab:entropy-results}
\centering
\scriptsize
\begin{tabular}{llrrrr}
\toprule
\textbf{Outcome} & \textbf{Predictor} & $\boldsymbol{\beta}$ & \shortstack{\textbf{$t$-summary}\\\textbf{95\% interval}} & \textbf{Exact $p$} & \textbf{BH $q$} \\
\midrule
Sample-sequence first-order entropy rate & Mean surprisal & $-.0063$ & [$-.0420$, .0295] & .7245 & .9622 \\
 & Summed surprisal & $-.0104$ & [$-.0665$, .0457] & .7047 & .9622 \\
 & Parse depth & .0149 & [$-.0219$, .0517] & .4012 & .9061 \\
 & Flesch--Kincaid grade & .0101 & [$-.0465$, .0668] & .7190 & .9622 \\
 & Mean dependency length & .0169 & [$-.0511$, .0848] & .6141 & .9622 \\
\addlinespace
Jump-sequence first-order entropy rate & Mean surprisal & $-.0177$ & [$-.0537$, .0183] & .3318 & .8974 \\
 & Summed surprisal & $-.0068$ & [$-.0701$, .0564] & .8221 & .9622 \\
 & Parse depth & $-.0133$ & [$-.0474$, .0208] & .4168 & .9061 \\
 & Flesch--Kincaid grade & .0015 & [$-.0537$, .0567] & .9687 & .9951 \\
 & Mean dependency length & .0210 & [$-.0234$, .0653] & .3311 & .8974 \\
\bottomrule
\end{tabular}
\begin{flushleft}\footnotesize Controlled equal-participant standardized slopes. Intervals are descriptive participant-$t$ compatibility summaries, not inversions of the exact test. Exact two-sided sign-flip $p$-values are shown; BH correction covers all 100 baseline-by-predictor tests, including these ten entropy rows.\end{flushleft}
\end{table*}

Table~\ref{tab:entropy-results} confirms that neither sample-sequence nor jump-sequence first-order entropy rate is associated with a linguistic predictor after BH correction across the 100 baseline--predictor tests. These comparators address sequence uncertainty and do not duplicate METER-Jump's reference-relative destination divergence.

\begin{table*}[htpb]
\caption{Empirical Performance of the Transition-Confidence Filter}
\label{tab:confidence-filter-audit}
\centering
\small
\begin{tabularx}{\textwidth}{Xccc}
\toprule
\textbf{Diagnostic} & \textbf{Median} & \textbf{IQR} & \textbf{5th--95th percentile} \\
\midrule
Participant confidence threshold & .00379 & .00333--.00421 & .00273--.00495 \\
Analysis-target jumps removed (\%) & 9.62 & 6.67--12.82 & 2.44--19.07 \\
Calibration-bank jumps retained (\%) & 89.99 & 89.99--90.00 & 89.99--90.00 \\
\midrule
\multicolumn{4}{l}{\textit{Aggregate target retention and composition by source state}} \\
State 1: retained / share of retained jumps (\%) & \multicolumn{3}{c}{90.00 / 31.35} \\
State 2: retained / share of retained jumps (\%) & \multicolumn{3}{c}{89.74 / 39.82} \\
State 3: retained / share of retained jumps (\%) & \multicolumn{3}{c}{90.30 / 28.83} \\
\bottomrule
\end{tabularx}
\begin{flushleft}\footnotesize The participant threshold is the calibration-bank 10th percentile of the two-endpoint assignment margin. Event summaries cover 4,175 primary analysis targets; bank retention is aggregated within participant before summarization. Across the six directed source--destination edges, exclusion ranged from 8.97\% to 10.50\%. The analysis was reconstructed from saved labels and primary target identities and used no linguistic outcome.\end{flushleft}
\end{table*}

Table~\ref{tab:confidence-filter-audit} shows that the calibration-derived threshold removes approximately the intended fraction of jumps without concentrating exclusions in one source state or directed edge. It supports the implementation check but does not establish that hard thresholding fully represents assignment uncertainty.

\begin{table}[htpb]
\caption{Decomposition of 1--40-Hz Versus 2--20-Hz METER-Jump Agreement}
\label{tab:processing-decomposition}
\centering
\small
\begin{tabular}{lccc}
\toprule
\textbf{Component} & \textbf{Global $\rho$} & \textbf{Median $\rho_u$} & \textbf{Median absolute difference} \\
\midrule
Observed $R_{\mathrm{jump}}$ & .637 & .602 & .00644 \\
Matched-null mean $\mu_0$ & .945 & 1.000 & .00239 \\
Matched-null SD $\sigma_0$ & .772 & .796 & .00265 \\
Centered $R_{\mathrm{jump}}-\mu_0$ & .384 & .283 & .00679 \\
Standardized $Z$ & .330 & .309 & .63410 \\
\bottomrule
\end{tabular}
\begin{flushleft}\footnotesize Spearman agreement over 4,167 common event identities. Seventeen of 18 participants retained the identical 96-event calibration bank and target-specific exclusion. Excluding the one changed-bank participant gave $\rho_Z=.318$. Component magnitudes have different units, so absolute differences should not be compared across rows.\end{flushleft}
\end{table}

Table~\ref{tab:processing-decomposition} localizes the low 1--40-Hz versus 2--20-Hz agreement. The raw divergence is only moderately concordant, matched-null centering reduces agreement further, and standardization yields the final $\rho_Z=.330$; the processing effect is therefore not explained by null-scale instability alone.

\begin{table*}[htpb]
\caption{Estimator-selection summaries obtained without linguistic outcomes}
\label{tab:estimator-selection}
\centering
\scriptsize
\setlength{\tabcolsep}{2pt}
\begin{tabular}{llcccc}
\toprule
\multicolumn{6}{l}{\textit{A. Transition shrinkage and concentration}}\\
\textbf{Shrinkage} & $\boldsymbol{\tau_A}$ & \shortstack{\textbf{Null FPR}\\\textbf{range}} & \shortstack{\textbf{Fixed-map}\\\textbf{detection}} & \textbf{Coverage} & \shortstack{\textbf{Coverage}\\\textbf{eligible}}\\
\midrule
Fixed & 1 & .029--.041 & .653 & .909 & Yes\\
Fixed & 4 & .035--.042 & .702 & .808 & No\\
Fixed & 16 & .032--.045 & .705 & .148 & No\\
State-adaptive & 1 & .034--.037 & .649 & .908 & Yes\\
State-adaptive & 4 & .036--.045 & .695 & .807 & No\\
State-adaptive & 16 & .033--.042 & .711 & .146 & No\\
\midrule
\multicolumn{6}{l}{\textit{B. Assignment-confidence policy at fixed $\tau_A=1$}}\\
\textbf{Policy} & & \shortstack{\textbf{Event-null FPR}\\\textbf{[95\% CI]}} & \shortstack{\textbf{Map-perturbed}\\\textbf{detection}} & \shortstack{\textbf{Mean retained}\\\textbf{jumps}} & \\
\midrule
Continuous weighting & & .041 [.030,.055] & .117 & 37.48 & \\
Drop lowest decile & & .036 [.026,.049] & .135 & 29.88 & \\
Hard assignment & & .034 [.024,.047] & .067 & 38.75 & \\
\bottomrule
\end{tabular}
\begin{flushleft}\footnotesize Results are from the numerical selection stream and precede the separately seeded two-seed evaluation. FPR ranges in panel A span the four evaluated null mechanisms. ``Eligible coverage'' required at least .90 coverage under the prespecified selection rule. The final choice used fixed $\tau_A=1$ because stronger concentration sharply reduced interval coverage; the lowest-decile policy gave the greatest map-perturbed detection among the confidence policies. Sparse-row prevalence was shared across paired policies because the simulated targets were identical; all estimators remained defined for empty rows. This grid compares policy families, not 0/5/10/20-percentile cutoffs; those threshold sensitivities remain unevaluated end to end.\end{flushleft}
\end{table*}

\begin{table*}[htpb]
\caption{Outcome-model recovery across 8,000 of the 16,000 simulated group fits}
\label{tab:outcome-recovery}
\centering
\scriptsize
\begin{tabular}{lrrrrr}
\toprule
\textbf{Score treatment} & \textbf{True $\beta$} & \textbf{Bias} & \textbf{RMSE} & \textbf{Coverage} & \textbf{Rejection / power}\\
\midrule
Single noisy score & 0.0 & .0017 & .0375 & .950 & .050\\
Single noisy score & 0.1 & $-.0290$ & .0480 & .886 & .414\\
Single noisy score & 0.2 & $-.0596$ & .0705 & .698 & .927\\
Single noisy score & 0.3 & $-.0891$ & .0975 & .410 & 1.000\\
\midrule
Integrated, 32 draws & 0.0 & .0012 & .0486 & .941 & .059\\
Integrated, 32 draws & 0.1 & $-.00001$ & .0460 & .961 & .521\\
Integrated, 32 draws & 0.2 & $-.0006$ & .0482 & .951 & .973\\
Integrated, 32 draws & 0.3 & $-.0049$ & .0478 & .947 & 1.000\\
\bottomrule
\end{tabular}
\\[8pt]
\textbf{Complete design-preserving group-null check}\\[3pt]
\begin{tabular}{lcccc}
\toprule
\textbf{Resampling unit} & \textbf{Replicates} & \textbf{Predictors} & \textbf{FWER} & \textbf{95\% Wilson interval} \\
\midrule
Within-participant sentence-row permutation & 500 & 5 & .054 & [.037, .077] \\
\bottomrule
\end{tabular}
\begin{flushleft}\footnotesize Each recovery row contains 1,000 fits with 18 participants. The displayed comparison isolates score-noise treatment. The remaining 8,000 fits comprised balanced and participant-imbalance scenarios; their null rejection rates were .052 and .048, null-parameter coverage ranged from .948 to .952, and coverage across individual nonzero-effect cells ranged from .946 to .963. The artificial single-score condition adds score uncertainty and is not the empirical METER point estimator. The group-null check preserved each participant's complete covariate design and 32 score draws, jointly permuted sentence rows within participant, recomputed the five integrated slopes, and applied the same exact $2^{18}$ sign-flip max-$T$(5) rule.\end{flushleft}
\end{table*}

Tables~\ref{tab:estimator-selection} and \ref{tab:outcome-recovery} make two previously dispersed decisions visible: $\tau_A=1$ was the only concentration with acceptable interval coverage, integration over score draws removed the attenuation and undercoverage induced by a deliberately noisy one-score analysis, and the complete group procedure controlled family-wise error under the design-preserving null.

The component study is paired at the level of the simulated participant, calibration bank, target event, and null regime. The uncalibrated variants use cutoffs learned only from the first half of the participant-fitted null stream; no cutoff is retuned for unequal occupancy, rare transitions, or map uncertainty. METER needs no global learned cutoff, so all 1,000 of its null ranks remain valid tests. Power comparisons use the common 500-replicate evaluation half. This distinction explains the different null sample sizes in Table~\ref{tab:component-ablation} without giving METER access to application outcomes.

The reference comparison addresses a specific estimand. Its non-null condition shifts every analysis event while leaving the predesignated target-disjoint bank unchanged. It therefore asks whether a complete measurement formulation can detect a shared shift relative to that fixed simulation baseline. LOO and balanced two-fold scores instead compare each event with other shifted events, so their attenuation is expected rather than an implementation defect. Reference composition and calibration differ across formulations; this is not an equal-reference estimator ablation. The experiment supports a fixed-bank contrast for this shared-shift question but does not imply that target-disjoint references dominate cross-fitting for every prediction or estimation problem.

\subsection{Exploratory dwell implementation}
\label{sec:dwell-details}

The censor-aware dwell coordinate uses the millisecond bin edges
\begin{equation*}
\begin{split}
[0,4,8,16,32,64,128,\;&256,512,1024,\;2048,\infty).
\end{split}
\end{equation*}
A complete episode contributes to one bin, whereas an episode extending beyond sentence offset contributes the survival probability above its observed lower-bound bin. Reference bin probabilities receive a Jeffreys mass of $1/2$ per bin, and target probabilities shrink toward a sampled reference distribution with $\tau_F=4$. Reference chains use 512 retained Gibbs draws after 128 burn-in iterations. Target calculations retain at least 512 draws (1,999 in the final run) after at most 64 burn-in iterations, with 16 conditional updates when the prior varies across reference draws. Divergence uses the same reference source-state weights as the jump coordinate, and each pseudo-event uses 128 inner draws under the symmetric $N-1$ construction. These details support reproduction of the exploratory quantity; they do not elevate dwell to a primary outcome.

The main manuscript's implementation table consolidates the study choices and their scientific roles. The detailed algorithms and sensitivity analyses here provide the information needed to reproduce and qualify those choices without duplicating the summary table.

Table~\ref{tab:rank-randomization} separates the deterministic bounds of the event rank from the stored randomized realization; Table~\ref{tab:bank-sensitivity} summarizes dependence on the balanced bank allocation. The first check concerns application-event rejection counts, whereas the group association analysis uses $Z$ and is unchanged.

\begin{center}
\captionof{table}{Sensitivity of application-event rejection counts to randomized rank position}
\label{tab:rank-randomization}
\centering
\small
\begin{tabular}{lcc}
\toprule
\textbf{Rank summary} & \textbf{Rejected events / 4,175} & \textbf{Fraction} \\
\midrule
Attainable lower endpoint & 312 & .0747 \\
Stored hash-randomized rank & 198 & .0474 \\
Expected over independent rank randomizers & 198 & .0474 \\
Central 95\% aggregate range over randomizers & 188--209 & .0450--.0501 \\
Conservative upper endpoint & 160 & .0383 \\
\bottomrule
\end{tabular}
\begin{flushleft}\footnotesize Each application event had 24 duration-matched pseudo-events, so one attainable rank interval had width $1/25=.04$. There were 160 events whose complete interval lay at or below .05 and 152 whose interval crossed .05. Under the implemented formula, each boundary event is rejected with probability .25; the aggregate range is the exact central 95\% binomial range for 152 such events. The stored SHA-256 randomizer produced exactly the expected total. Randomization can change which boundary events reject, but it does not enter the continuous $Z$ coordinate or the participant-level association analysis. The lower and upper endpoints are sensitivity bounds, not replacement primary tests.\end{flushleft}
\end{center}

\begin{center}
\captionof{table}{Sensitivity to 20 independently balanced calibration banks}
\label{tab:bank-sensitivity}
\centering
\scriptsize
\begin{tabular}{lcc}
\toprule
\textbf{Diagnostic for alternative banks versus bank 0} & \textbf{Median} & \textbf{Range} \\
\midrule
Overlapping-event METER-Jump $Z$ rank correlation & .9379 & .9309--.9462 \\
Participant-slope correlation over five predictors & .6482 & .5571--.7555 \\
Pseudo-event rejection rate & .0492 & .0446--.0527 \\
Maximum $|\Delta\beta|$ across five group slopes & .0367 & .0166--.0528 \\
Minimum within-bank max-$T$ $p$ & .6083 & .1014--.7989 \\
\bottomrule
\end{tabular}
\begin{flushleft}\footnotesize Bank 0 is the primary bank. Repeats 1--19 used the same joint balancing rule; target overlap was 2,985--3,042. Calibration remained nominal and no corrected conclusion changed. Lower slope agreement shows that bank composition affected effect magnitude more than event ordering.\end{flushleft}
\end{center}